\documentclass[10pt,journal,compsoc]{IEEEtran-customize}

\ifCLASSOPTIONcompsoc
  \usepackage[nocompress]{cite}
\else
  \usepackage{cite}
\fi
\usepackage{graphicx}
\usepackage{color}
\usepackage{xcolor}
\usepackage{booktabs}
\usepackage{multirow}
\usepackage[hyphens]{url}
\usepackage{hyperref}
\usepackage{amsmath}
\usepackage{amsfonts, amssymb}
\usepackage{colortbl}
\usepackage{subfigure}

\definecolor{purple}{RGB}{128, 0, 128}
\definecolor{LightRed}{rgb}{1,0.92,0.92}
\definecolor{LightOrange}{rgb}{1,0.95,0.88}
\definecolor{LightYellow}{rgb}{1.0,1.0,0.84}
\definecolor{LightGreen}{rgb}{0.9,1.0,0.88}
\definecolor{LightCyan}{rgb}{0.9,1,1}
\definecolor{LightBlue}{rgb}{0.9,0.94,1}
\definecolor{LightIndigo}{rgb}{0.92,0.9,1}
\definecolor{LightMagenta}{rgb}{0.96,0.86,1}
\definecolor{DirtyWhite}{rgb}{0.96,0.96,0.96}

\DeclareSymbolFont{extraup}{U}{zavm}{m}{n}
\DeclareMathSymbol{\varheart}{\mathalpha}{extraup}{86}
\DeclareMathSymbol{\vardiamond}{\mathalpha}{extraup}{87}
\DeclareMathSymbol{\varclubsuit}{\mathalpha}{extraup}{88}

\begin{document}
\title{A Systematic Review of Deep Learning-based Research on Radiology Report Generation}

\author{
        Chang Liu$^{\spadesuit}$,~
        Yuanhe Tian$^{\spadesuit\varheart}$,~
        and ~Yan Song$^{\spadesuit*}$ \\
    \vspace{0.2cm}

    $^{\spadesuit}$University of Science and Technology of China \hspace{0.1cm}
    $^{\varheart}$University of Washington
    \\
    
    \vspace{0.1cm}
    
    $^{\spadesuit}$\texttt{lc980413@mail.ustc.edu.com} \hspace{0.1cm}
    $^{\varheart}$\texttt{yhtian@uw.edu} \hspace{0.1cm}
    $^{\spadesuit}$\texttt{clksong@gmail.com}  \\
}


\IEEEtitleabstractindextext{%
\begin{abstract}

Radiology report generation (RRG) aims to automatically generate free-text descriptions from clinical radiographs, e.g., chest X-Ray images.
RRG plays an essential role in promoting clinical automation and presents significant help to provide practical assistance for inexperienced doctors and alleviate radiologists' workloads.
Therefore, consider these meaningful potentials, research on RRG is experiencing explosive growth in the past half-decade, especially with the rapid development of deep learning approaches.
Existing studies perform RRG from the perspective of enhancing different modalities, provide insights on optimizing the report generation process with elaborated features from both visual and textual information,
and further facilitate RRG with the cross-modal interactions among them.
In this paper, we present a comprehensive review of deep learning-based RRG from various perspectives.
Specifically, we firstly cover pivotal RRG approaches based on the task-specific features of radiographs, reports, and the cross-modal relations between them,
and then illustrate the benchmark datasets conventionally used for this task with evaluation metrics, subsequently analyze the performance of different approaches and finally offer our summary on the challenges and the trends in future directions.
Overall, the goal of this paper is to serve as a tool for understanding existing literature and inspiring potential valuable research in the field of RRG.$^\dagger$

\end{abstract}

}

\maketitle
\IEEEdisplaynontitleabstractindextext
\IEEEpeerreviewmaketitle

\makeatletter
\def\@IEEEcompsocmakefnmark{\hbox{\normalfont\@thefnmark\ }}
\long\def\@makefntext#1{\parindent 1em\indent\hbox{\@IEEEcompsocmakefnmark}#1}
\makeatother

\def\thefootnote{*}\footnotetext{Corresponding author.}
\def\thefootnote{\textdagger}\footnotetext{We maintain a GitHub project at \url{https://github.com/synlp/RRG-Review} to summarize RRG-related papers and resources, which is under continuously updating to facilitate future research in this field.}
\makeatletter
\def\@IEEEcompsocmakefnmark{\hbox{\normalfont\@thefnmark.\ }}
\long\def\@makefntext#1{\parindent 1em\indent\hbox{\@IEEEcompsocmakefnmark}#1}
\makeatother

\renewcommand{\thefootnote}{\arabic{footnote}}

\IEEEraisesectionheading{\section{Introduction}\label{sec:introduction}}

\IEEEPARstart{W}{ith} the prosperity of artificial intelligence (AI), medical industry has emerged a growing demand for its advanced techniques, where different medical applications in urgent needs of automatically processing massive medical data.
These applications usually comprise multiple research directions, including pathological image analysis, medical report generation, disease prediction and diagnosis, etc.
In performing the aforementioned research, AI techniques are expected to handle medical data in various modalities.
Particularly, medical imaging (e.g., X-Ray, MRI) serves as the mainstream medical data that are collected by different medical instruments, which presents the patients' health condition straightforwardly through the various image types.
Besides image, text is another important medium playing a pivotal carrier in doctors' daily routines and medical researches, which stores various medical contents through the formulation of natural language, e.g., medical records, research papers, reports.
Both image and text synergistically perform in clinical diagnosis and treatment recommendation, where, particularly, medical imaging usually requires physicians to write reports according to the syndromes of patients that are reflected in the image so as to form professional records for later processing.
Among all applications requiring such connection between image and text,
automatically performing radiology report generation (RRG) serves as the most practical and extensive support in the imaging department of many hospitals for lightening radiologists' burdens and assisting inexperienced physicians.
%
Technically,
RRG aims to generate descriptive texts for a radiograph, 
which requires the produced texts to accurately depict overall and local health condition (e.g., regional lesions, disease of specific organs), meanwhile maintaining the consistency of the entire report to the input image thus formulating the final diagnosis.
Thus this task holds great significance of applying AI techniques to the medical industry, so that it emerges as an attractive research direction in AI and clinical medicine in recent years.

\begin{figure}[t]
    \centering
    \includegraphics[width=1.0\linewidth, trim=0 10 0 0]{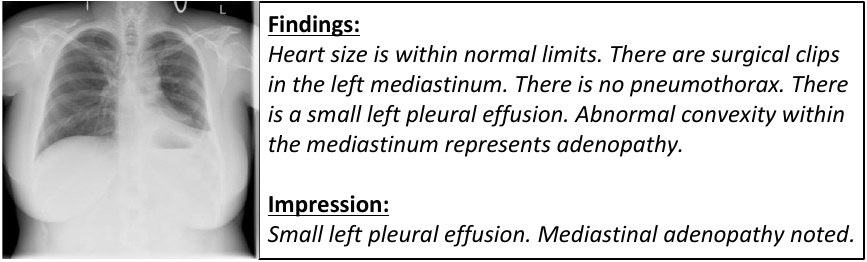}
    \caption{A representative chest radiology image with its corresponding doctor-written radiology report.
    Specifically, each report consists of a ``Findings'' and a ``Impression'' section, where the ``Findings'' section records detailed descriptions by radiologists, and the ``Impression'' section presents an overall diagnostic summarization based on the radiograph.
    RRG is normally aiming at generating the ``Findings'' content.
    }
    \label{fig: tissor}
    \vskip -1.0em
    \end{figure}

To better understand RRG,
Figure \ref{fig: tissor} presents an example of a chest radiology image with its description including doctor-written findings and impression.
The findings section consists of detailed descriptions recorded by radiologists for each specific region of the radiology image, while the impression section contains the most information summarized from the findings section for the final clinical diagnosis.
Normally RRG is targeted to generate the findings section,
i.e., radiology report, according to an input radiograph.
To perform RRG, early approaches  produce medical labels (e.g., locations of lesions and severities) to assist the generation process,
%
where such approaches consist of two main categories, i.e., retrieval-based \cite{retrieval-1, retrieval-2} and disease classification-based approaches \cite{classification-1, classification-2}, respectively.
Particularly, the former category aims to correlate the visual features of radiographs to specific texts (i.e, disease attributes) in the report, and obtain relevant texts through retrieving a medical database.
The latter category manages to directly predict medical labels from the input radiograph.
For both categories of the approaches, however, they still rely on professional radiologists to complete the writing of the entire report thus fail to achieve full automation in practical clinic environment, which drives the expectation of
%
more effective approaches to handle this task.

Recently,
with the rising of deep learning and neural networks, significant breakthrough has been witnessed in RRG research, where studies have dramatically grown in the past five years, especially with the trend presenting an explosive curve in the recent two years.
Upon training on large-scale radiograph-report pairs in prevailing radiology datasets, neural approaches \cite{jing-etal-2018-automatic, nips-2018-hrgr-agent, nishino-etal-2020-reinforcement, chen-etal-2020-generating, nooralahzadeh-etal-2021-progressive-transformer, you-etal-2022-jpg, delbrouck-etal-2022-improving, cvpr-2023-metransformer, cvpr-2023-kiut} are capable of generating the entire contents of radiology reports in an end-to-end manner.
These approaches are mostly based on the encoder-decoder architecture, which normally employ visual encoders to extract visual representations from radiographs, and adopt text decoders to generate descriptive texts of radiology reports according to the extracted representations.
%
Moreover,
they also further consider task-specific features of RRG from the aspects of different modalities \cite{jing-etal-2019-show, aaai-2019-knowledge, aaai-2020-when, cvpr-2021-exploring, chen-etal-2021-cross-modal, iccv-2021-visual-textual, you-etal-2022-jpg, nishino-etal-2022-factual, qin-song-2022-reinforced, kale-etal-2023-replace, cvpr-2023-interactive, cov-ctr}.
Specifically, some of them emphasize on properties of radiographs or reports separately \cite{aaai-2019-knowledge, jing-etal-2019-show, aaai-2020-when, cvpr-2021-exploring, nishino-etal-2022-factual, kale-etal-2023-replace, cvpr-2023-interactive, cov-ctr, kale-etal-2023-kgvl, hou-etal-2023-organ},
while others focus on establishing the vision-language connections between radiographs and reports \cite{yan-etal-2021-weakly-supervised, chen-etal-2021-cross-modal, cvpr-2021-self-boosting, liu-etal-2021-contrastive, iccv-2021-visual-textual, qin-song-2022-reinforced, miura-etal-2021-improving}.
All these approaches have achieved great success on the RRG task, and provide interesting findings, guidance as well as references to push this research area forward.

In this paper, we present an up-to-date and comprehensive overview of deep learning-based RRG, with particular emphases on different algorithms, datasets, and evaluation standards.
We start from introducing background knowledge of the RRG task, and demonstrate our classification criteria to categorize existing deep learning-based RRG approaches based on the perspectives of enhancing corresponding modalities, including visual-only, textual-only, as well as cross-modal approaches.
%
Then we present prevailing RRG datasets in multiple domains,
including benchmarks for standard evaluation for RRG, and domain-specific datasets that emphasize on radiology studies of particular organs (e.g., CT imaging of COVID patients, liver lesions), etc.
%
Afterwards, we summarize the evaluation principles
and a series of metrics that measure the performance of RRG from different perspectives.
Subsequently we analyze experimental results from different approaches based on the aforementioned evaluation metrics,
%
and analyze the impacts of different network architectures on
RRG performance.
Furthermore, we discuss the challenges and trends in future directions to extend current RRG research,
from perspectives of model design, modality enhancement, data augmentation, as well as evaluation metrics.

\section{Approaches} \label{sec: approaches}

RRG in general follows a cross-modal generation paradigm similar to image captioning tasks \cite{iclr2015_image_captioning, rennie2017selfcritical, lu2017knowing, anderson2018bottomup, jing-etal-2019-show, cornia2020m2}, yet different in the resulting text length.
Formally,
given a radiology image $\mathcal{I}$, RRG generates its corresponding radiology report $\widehat{\mathcal{R}}$, with distilling key semantic information from $\mathcal{I}$, and accurately produce descriptive texts in $\widehat{\mathcal{R}}$.
Existing approaches generally adopt the encoder-decoder architecture as their foundation pipeline, which employs a visual encoder $f_v$ to extract high-level semantics (e.g., latent representations, medical terms, or semantic graphs) that are represented by $\mathcal{H}^v$ from $\mathcal{I}$, and utilizes a text decoder $f_t$ to transfer $\mathcal{H}^v$ into descriptive texts in $\widehat{\mathcal{R}}$, with the overall process formulated by
\begin{equation} \label{eq: standard_rrg_paradigm}
      \widehat{\mathcal{R}} = f_t\left(\mathcal{H}^v\right), \ \ \mathcal{H}^v = f_v\left(\mathcal{I}\right)
\end{equation}
It is worth noting that most existing approaches generate $\widehat{\mathcal{R}}$ based on a single radiograph $\mathcal{I}$; 
although there are some approaches that process multiple radiographs with support from particular datasets, they use the concatenation of these radiographs to form them into a single input image and thus fit into our task formulation.
In learning the aforementioned pipeline,
existing studies utilize radiology reports $\mathcal{R}^*$ written by professional radiologists as the reference for their systems, whose
output $\widehat{\mathcal{R}}$ is expected to be as close to $\mathcal{R}^*$ as possible.
Particularly, they
follows standard optimization procedure by generative neural models by utilizing the cross-entropy loss $\mathcal{L}_{CE}$ through comparing system output $\widehat{\mathcal{R}}$ with the gold standard report $\mathcal{R}^*$.
In detail,
let $\widehat{\mathcal{R}} = \{ \widehat{r}_{1} \dots \widehat{r}_{N_r} \}$ and $\mathcal{R}^* = \{ r^*_1 \dots r^*_{N_{r^*}} \}$ with $N_r$ and $N_{r^*}$ denoting text token numbers in $\widehat{\mathcal{R}}$ and $\mathcal{R}^*$, respectively.
%
To generate the $i$-th token, the RRG model takes 
the radiograph $\mathcal{I}$ and historical generated tokens (which are $r^*_1 \dots r^*_{i-1}$ in training and $\widehat{r}_{1} \dots \widehat{r}_{i-1}$ in inference) and computes the probability distribution over the vocabulary $\mathcal{V}$ where the probability of generating the $j$-th token $v_j$ in the vocabulary is denoted as $p_{i,j}$.
Then, RRG model selects the token with the highest probability as the generated token $\widehat{r}_{i} = arg\,max_{v_j \in \mathcal{V}} (p_{i,j})$.
Afterwards, $\widehat{r}_i$ is compared with $r^*_{i}$ to calculate the cross-entropy loss $\mathcal{L}^i_{CE}$ of the $i$-th token by
\begin{equation} \label{eq: cross-entropy}
    \mathcal{L}^i_{CE} = -\sum_{v_j \in \mathcal{V}} p^*_{i,j} \log p_{i,j} 
\end{equation}
where $p^*_{i,j}$ is a probability defined by
$p^*_{i, j}=1$ if $v_j=r^*_i$ and $p^*_{i,j}=0$ otherwise.
%
Thus, $\mathcal{L}_{CE}$ over all tokens of $\widehat{\mathcal{R}}$ is obtained through $\mathcal{L}_{CE} = \sum^{N_r}_{i=1} \mathcal{L}^i_{CE}$.

\definecolor{lightgray}{gray}{0.97}
\begin{table*}[t]
\centering
\caption{
    Categorization of existing RRG approaches based on our classification criteria of enhancing different modalities, including visual-only approaches, textual-only approaches, and cross-modal approaches, which are highlighted in red, green, and blue background, respectively.
    Additionally, we list the model architecture and evaluation standard of the surveyed studies.
    ``GCN'' denotes graph convolutional networks.
    }
\label{tab: approach_classification}
\vspace{-0.2cm}
\rowcolors{4}{lightgray}{}
\resizebox{\linewidth}{!}{
\begin{tabular}{lcccccc}
\toprule
 & & \multicolumn{2}{c}{\textbf{Model Architecture}} & & \multicolumn{2}{c}{\textbf{Evaluation Standard}} \\
\cmidrule{3-4} \cmidrule{6-7}
\textbf{Study} & \textbf{Method Category} & Encoder & Decoder & & Datasets & Evaluation Metrics \\
\midrule
\rowcolor{LightRed}
Wang \textit{et al.} \cite{Wang2022SelfAG} & visual-only & Transformer & Transformer && IU X-Ray, MIMIC-CXR & NLG metrics \\
\rowcolor{LightRed}
Tanida \textit{et al.} \cite{cvpr-2023-interactive} & visual-only & CNN & Transformer && MIMIC-CXR & NLG metrics, CE metrics \\
\rowcolor{LightRed}
Li \textit{et al.} \cite{cov-ctr} & visual-only & GCN & Transformer && CX-CHR, COV-CTR & NLG metrics, CIDEr \\
\midrule
\rowcolor{LightGreen}
Xue \textit{et al.} \cite{xue2018multimodal} & textual-only & CNN & LSTM && IU X-Ray & NLG metrics \\
\rowcolor{LightGreen}
Yuan \textit{et al.} \cite{yuan2019automatic} & textual-only & CNN & LSTM && IU X-Ray & NLG metrics \\
\rowcolor{LightGreen}
Jing \textit{et al.} \cite{jing-etal-2019-show} & textual-only & LSTM & LSTM && IU X-Ray, CX-CHR & NLG metrics, CIDEr \\
\rowcolor{LightGreen}
Li \textit{et al.} \cite{aaai-2019-knowledge} & textual-only & Transformer & Transformer && IU X-Ray, CX-CHR & NLG metrics, CIDEr \\
\rowcolor{LightGreen}
Harzig \textit{et al.} \cite{harzig2019addressing} & textual-only & CNN & LSTM && IU X-Ray & NLG metrics \\
\rowcolor{LightGreen}
Zhang \textit{et al.} \cite{aaai-2020-when} & textual-only & GCN & LSTM && IU X-Ray, CX-CHR & NLG metrics, CIDEr, MIRQI \\
\rowcolor{LightGreen}
Chen \textit{et al.} \cite{chen-etal-2020-generating} & textual-only & Transformer & Transformer && IU X-Ray, MIMIC-CXR & NLG metrics, CE Metrics \\
\rowcolor{LightGreen}
Liu \textit{et al.} \cite{cvpr-2021-exploring} & textual-only & Transformer & Transformer && IU X-Ray, MIMIC-CXR & NLG metrics \\
\rowcolor{LightGreen}
Yan \textit{et al.} \cite{yan-etal-2021-weakly-supervised} & textual-only & Transformer & Transformer && MIMIC-ABN, MIMIC-CXR & NLG metrics, CE Metrics \\
\rowcolor{LightGreen}
Liu \textit{et al.} \cite{liu2021auto} & textual-only & Transformer & Transformer && IU X-Ray, MIMIC-CXR & NLG metrics \\
\rowcolor{LightGreen}
Nooralahzadeh \textit{et al.} \cite{nooralahzadeh-etal-2021-progressive-transformer} & textual-only & Transformer & Transformer && IU X-Ray, MIMIC-CXR & NLG metrics, CE Metrics \\
\rowcolor{LightGreen}
You \textit{et al.} \cite{you-etal-2022-jpg} & textual-only & Transformer & Transformer && IU X-Ray & NLG metrics \\
\rowcolor{LightGreen}
Nishino \textit{et al.} \cite{nishino-etal-2022-factual} & textual-only & CNN & Transformer && JLiverCT, MIMIC-CXR & NLG metrics, CE metrics, CO \\
\rowcolor{LightGreen}
Dalla Serra \textit{et al.} \cite{dalla-serra-etal-2022-multimodal} & textual-only & Transformer & Transformer && MIMIC-CXR & NLG metrics, CE metrics \\
\rowcolor{LightGreen}
Kale \textit{et al.} \cite{kale-etal-2023-kgvl} & textual-only & GCN & Transformer && IU X-Ray, MIMIC-CXR & NLG metrics, BERTScore \\
\rowcolor{LightGreen}
Yan \textit{et al.} \cite{yan2023attributed} & textual-only & CNN & Transformer && IU X-Ray, MIMIC-CXR & NLG metrics, CE Metrics \\
\rowcolor{LightGreen}
Li \textit{et al.} \cite{cvpr-2023-dynamic} & textual-only & Transformer & Transformer && IU X-Ray, MIMIC-CXR & NLG metrics, CE metrics, CIDEr \\
\rowcolor{LightGreen}
Hou \textit{et al.} \cite{hou-etal-2023-organ} & textual-only & Transformer & Transformer && IU X-Ray, MIMIC-CXR & NLG metrics, CE metrics \\
\rowcolor{LightGreen}
Kale \textit{et al.} \cite{kale-etal-2023-replace} & textual-only & CNN & Transformer && IU X-Ray, MIMIC-CXR & NLG metrics, CE metrics, CIDEr \\
\rowcolor{LightGreen}
Zhang \textit{et al.} \cite{TMM-semi-supervised} & textual-only & CNN & Transformer && IU X-Ray, MIMIC-CXR & NLG metrics, CE metrics \\
\midrule
\rowcolor{LightBlue}
Jing \textit{et al.} \cite{jing-etal-2018-automatic} & cross-modal & CNN & LSTM && IU X-Ray, PEIR Gross & NLG metrics \\
\rowcolor{LightBlue}
Nishino \textit{et al.} \cite{nishino-etal-2020-reinforcement} & cross-modal & RNN & GRU && CX-CHR & NLG metrics \\
\rowcolor{LightBlue}
Wang \textit{et al.} \cite{cvpr-2021-self-boosting} & cross-modal & CNN & LSTM && IU X-Ray, COV-CTR & NLG metrics, CIDEr \\
\rowcolor{LightBlue}
Liu \textit{et al.} \cite{liu-etal-2021-contrastive} & cross-modal & CNN & LSTM && IU X-Ray, MIMIC-CXR & NLG metrics, CE metrics \\
\rowcolor{LightBlue}
Ni \textit{et al.} \cite{iccv-2021-visual-textual} & cross-modal & CNN & LSTM && IU X-Ray, MIMIC-CXR & NLG metrics, CIDEr, nKTD \\
\rowcolor{LightBlue}
Chen \textit{et al.} \cite{chen-etal-2021-cross-modal} & cross-modal & Transformer & Transformer && IU X-Ray, MIMIC-CXR & NLG metrics, CE Metrics \\
\rowcolor{LightBlue}
Hou \textit{et al.} \cite{miccai-2021-hou-etal-ratchet} & cross-modal & CNN & Transformer && IU X-Ray, MIMIC-CXR & NLG metrics, CE metrics, CIDEr \\
\rowcolor{LightBlue}
You \textit{et al.} \cite{you2021aligntransformer} & cross-modal & Transformer & Transformer && IU X-Ray, MIMIC-CXR & NLG metrics \\
\rowcolor{LightBlue}
Qin \textit{et al.} \cite{qin-song-2022-reinforced} & cross-modal & Transformer & Transformer && IU X-Ray, MIMIC-CXR & NLG metrics, CE Metrics \\
\rowcolor{LightBlue}
Wang \textit{et al.} \cite{wang2022cross} & cross-modal & Transformer & Transformer && IU X-Ray, MIMIC-CXR & NLG metrics \\
\rowcolor{LightBlue}
Delbrouck \textit{et al.} \cite{delbrouck-etal-2022-improving} & cross-modal & CNN & Transformer && IU X-Ray, MIMIC-CXR & NLG metrics, CE metrics, CIDEr \\
\rowcolor{LightBlue}
Yu \textit{et al.} \cite{icbb-2022-yu-etal-clinically-coherent} & cross-modal & Transformer & Transformer && IU X-Ray, MIMIC-CXR & NLG metrics, CE metrics, CIDEr \\
\rowcolor{LightBlue}
Yan \textit{et al.} \cite{bhi-2022-yan-etal-prior-guided} & cross-modal & Transformer & Transformer && IU X-Ray, MIMIC-CXR & NLG metrics, CE metrics, CIDEr \\
\rowcolor{LightBlue}
Wang \textit{et al.} \cite{miccai-2022-wang-etal-inclusive} & cross-modal & CNN & Transformer && IU X-Ray, MIMIC-CXR & NLG metrics \\
\rowcolor{LightBlue}
Wang \textit{et al.} \cite{miccai-2022-wang-etal-semantic-assisted} & cross-modal & Transformer & Transformer && MIMIC-CXR & NLG metrics \\
\rowcolor{LightBlue}
Kong \textit{et al.} \cite{miccai-2022-kong-etal-transq} & cross-modal & Transformer & Transformer && IU X-Ray, MIMIC-CXR & NLG metrics, CE metrics \\
\rowcolor{LightBlue}
Tanwani \textit{et al.} \cite{miccai=2022-tanwani-etal-repsnet} & cross-modal & CNN & Transformer && IU X-Ray & NLG metrics \\
\rowcolor{LightBlue}
Wang \textit{et al.} \cite{wang-etal-2022-automated} & cross-modal & Transformer & Transformer && IU X-Ray, MIMIC-CXR & NLG metrics, CIDEr \\
\rowcolor{LightBlue}
Yang \textit{et al.} \cite{yang-etal-2023-joint} & cross-modal & Transformer & Transformer && IU X-Ray, MIMIC-CXR & NLG metrics, CE metrics, CIDEr \\
\rowcolor{LightBlue}
Huang \textit{et al.} \cite{cvpr-2023-kiut} & cross-modal & Transformer & Transformer && IU X-Ray, MIMIC-CXR & NLG metrics, CE metrics \\
\rowcolor{LightBlue}
Wang \textit{et al.} \cite{cvpr-2023-metransformer} & cross-modal & Transformer & Transformer && IU X-Ray, MIMIC-CXR & NLG metrics, CE metrics, CIDEr \\
\rowcolor{LightBlue}
Nicolson \textit{et al.} \cite{warm-starting} & cross-modal & CNN & LSTM && IU X-Ray, MIMIC-CXR & NLG metrics, CIDEr \\
\bottomrule
\end{tabular}
}
\vskip -1em
\end{table*}

Owing to the cross-modal characteristics of RRG,
existing approaches put emphasis on facilitating RRG with its task-specific features from different modalities, where they consider the features from radiographs, reports, as well as the cross-modal correlations between them.
Therefore, in this paper, we categorize existing approaches into three main streams based on the modality that they focus on, including visual-only, textual-only, and cross-modal approaches, respectively, where Table \ref{tab: approach_classification} presents such categorization on different approaches.

\begin{figure*}[t]
\centering
\subfigure[]{
\includegraphics[height=0.2\linewidth]{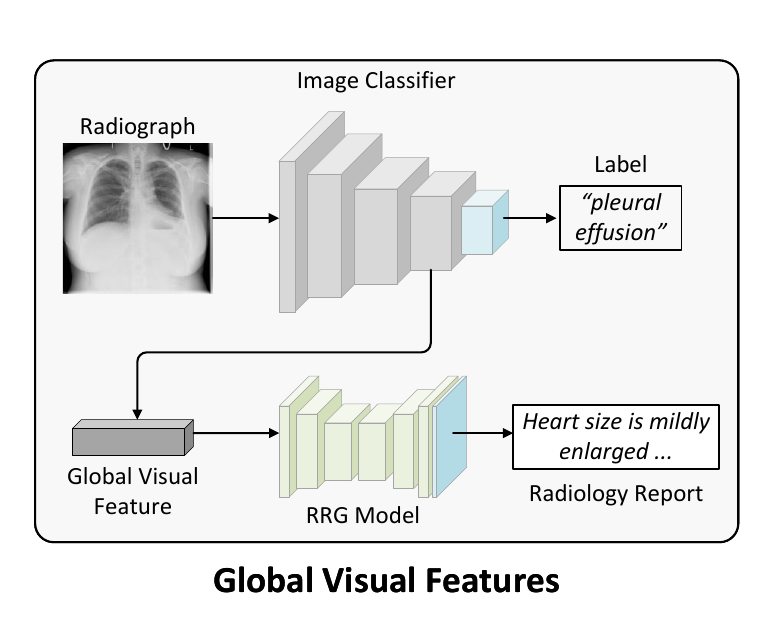}
\label{sfig: global-visual-feature}}
\subfigure[]{
\includegraphics[height=0.2\linewidth]{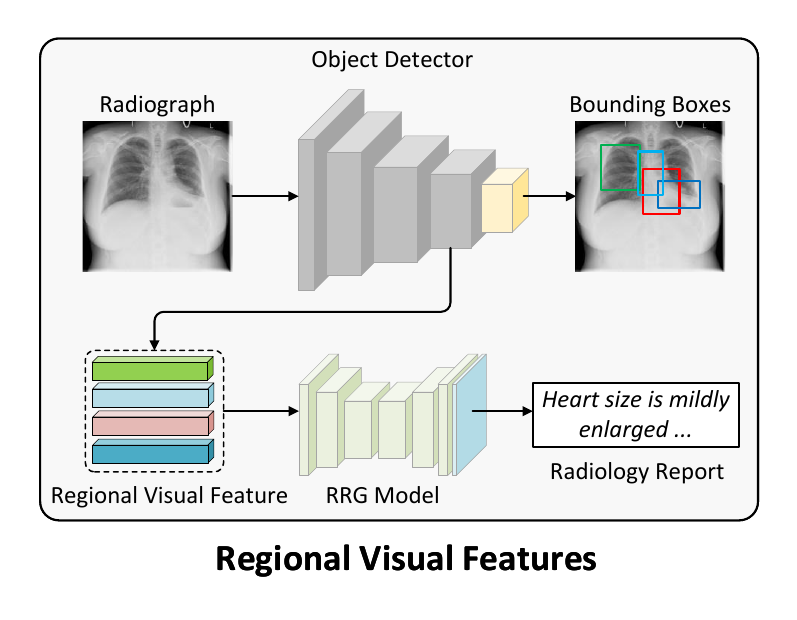}
\label{sfig: regional-visual-feature}}
\subfigure[]{
\includegraphics[height=0.2\linewidth]{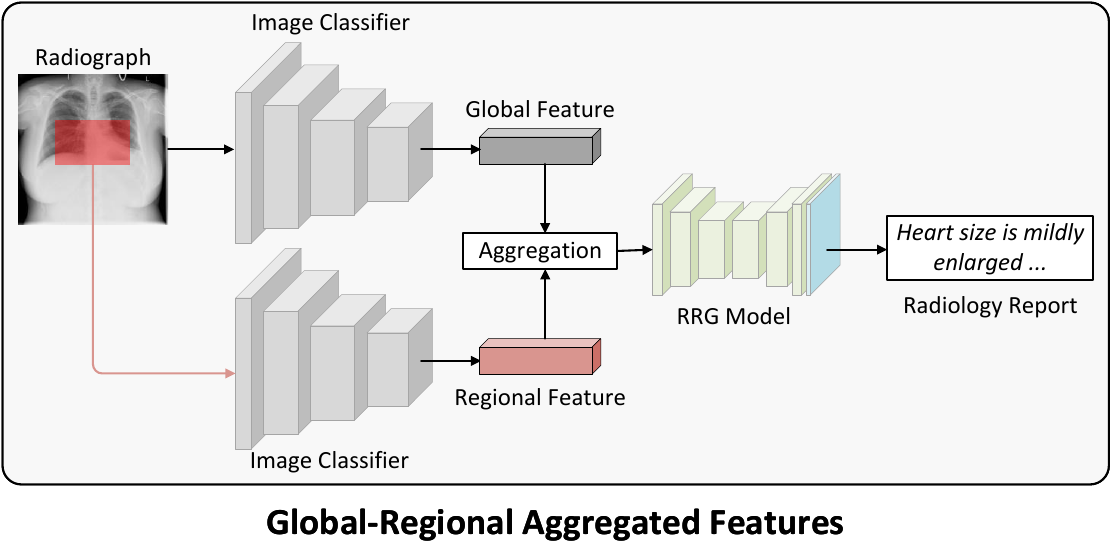}
\label{sfig: global-regional-aggregated-visual-feature}}
\vspace{-0.5cm}
\center
\caption{The architectures of three main categories of visual-only approaches that utilize different types of visual features for the report generation process, including (a) global visual features, (b) regional visual features, and (c) global-regional aggregated features.}
\vspace{-1em}
\label{fig: visual-only}
\end{figure*}

\subsection{Visual-only Approaches}
Serving as the input of RRG,
radiographs are generally the only information input to the task, therefore becomes vital on condition of their quality.
Specifically, they contains different granular features of RRG containing key insights that guide report writing in the following aspects.
First, a radiograph panoramically presents the overall health condition of a patient, which is highly associated with the consistency of the resulting report.
Second, particular regions of a radiograph indicate the status of a patient's specific organs, which determines the preciseness of describing particular disease attributes in the report.
Third, both global and regional features of a radiograph jointly provide necessary evidences for decision-making process for professional radiologists, especially when they write well-elaborated radiology reports.
Similar to the above analysis on different features,
visual-only approaches, which focus on processing radiographs,
are thus proposed to facilitate RRG through extracting different granular features.
%
In details, visual-only approaches generally consist of a visual feature extractor $f_{fe}$ and a visual feature encoder $f_{ve}$, where $f_{fe}$ is employed
to encode representative features $\mathcal{H}$ from $\mathcal{I}$ by
\begin{equation} \label{eq: visual_only_approach}
    \mathcal{H} = f_{fe} \left( \mathcal{I} \right)
\end{equation}
and $f_{ve}$ processes the resulted $\mathcal{H}$ to obtain the latent representation $\mathcal{H}^v$ through 
\begin{equation} \label{eq: visual-encoders}
    \mathcal{H}^v = f_{ve} \left(\mathcal{H} \right)
\end{equation}
Therefore, in the following texts,
we categorize visual-only approaches based on the granularity of $\mathcal{H}$, including global, regional, and global-regional aggregated visual features, respectively, whose architectures are presented in Figure \ref{fig: visual-only}.

\subsubsection{Global Visual Features}

%
Approaches in this category focus on the
global view of radiographs, where global visual feature is the most widely used feature type in existing studies.
Figure \ref{sfig: global-visual-feature} presents the visualization of approaches that adopt global visual features.
According to Eq. (\ref{eq: visual_only_approach}), there are approaches employ image classification models pre-trained on ImageNet \cite{deng2009imagenet} as $f_{fe}$, and fine-tune $f_{fe}$ with the training objective of RRG.
The global visual feature $\mathcal{H}$ is obtained from the last layer of $f_{fe}$ based on Eq. (\ref{eq: visual_only_approach}), and is further processed by $f_{ve}$ in Eq. (\ref{eq: visual-encoders}) into various formations that represent the high-level semantics of radiographs, e.g., latent representation \cite{jing-etal-2018-automatic, aaai-2019-knowledge, chen-etal-2020-generating, chen-etal-2021-cross-modal, qin-song-2022-reinforced, tanida2023interactive}, medical terms \cite{nishino-etal-2020-reinforcement, nooralahzadeh-etal-2021-progressive-transformer, nishino-etal-2022-factual, kale-etal-2023-replace}, and graph \cite{aaai-2020-when, cov-ctr, kale-etal-2023-kgvl}.
Particularly, early approach \cite{jing-etal-2018-automatic} utilizes pre-trained VGG-Net \cite{iclr-2015-vgg-net} for global visual feature extraction of radiographs, and subsequent studies \cite{jing-etal-2019-show, aaai-2020-when, chen-etal-2020-generating, chen-etal-2021-cross-modal, yan-etal-2021-weakly-supervised, cvpr-2021-exploring, you-etal-2022-jpg, qin-song-2022-reinforced, cov-ctr, kale-etal-2023-kgvl, hou-etal-2023-organ, kale-etal-2023-replace, cvpr-2023-kiut} manage to employ deeper and more complex classification models, e.g., ResNet \cite{he2016residual} and DenseNet \cite{huang2017densely}, to extract representative features from radiographs.
With the development of Transformer \cite{vaswani2017attention} architecture in the computer vision community, vision transformer (ViT) \cite{dosovitskiy2021an} demonstrated outstanding performance on image understanding, which is also adopted by recent RRG studies \cite{cvpr-2023-metransformer, cvpr-2023-dynamic} to extract more discriminative visual feature from radiographs.

\subsubsection{Regional Visual Features}

Although global visual features dominate visual-only approaches,
they are still ambiguous to represent specific regions of radiographs,
which potentially lead to inaccurate descriptions of particular disease attributes in the generated reports.
To provide more fine-grained visual information for RRG, some approaches \cite{cvpr-2021-self-boosting, cvpr-2023-interactive}  extract regional features radiographs.
Figure \ref{sfig: regional-visual-feature} presents the visualization of approaches that adopt regional visual features.
These approaches put emphasis on local pathological lesions in radiographs, and extract features from them to provide conditional guidance to describing particular disease attributes (e.g., abnormalities).
In doing so, they need to first obtain a series of anatomical regions $\{ \mathcal{I}_1 \dots \mathcal{I}_{N_a} \}$ in $N_a$ categories based on human annotations or object detection algorithms.
Then, they fine-tune pre-trained object detection models as $f_{fe}$ according to Eq. (\ref{eq: visual_only_approach}), and extract $\mathcal{H}$ from $\{ \mathcal{I}_1 \dots \mathcal{I}_{N_a} \}$.

%
To facilitate RRG with regional visual features, early approaches mainly utilize hand-crafted object detection algorithms to detect regions.
For example, Wang \textit{et al.} \cite{cvpr-2021-self-boosting} adopts a selective search algorithm \cite{ijcv-2013-selective-search} to generate regions for each radiograph in an unsupervised manner, and employ rules to select result regions from them.
%
The approach then
extracts features from the selected regions, and utilizes a self-attention layer to learn the internal relationships among the regions, results in relation-enhanced features for the subsequent report generation process.
However, in doing so, it faces a problem in effectively locating specific organs in radiographs since the regions detected by hand-crafted object detection algorithms (e.g., selective search algorithm) usually do not precise and overlap with each other.
%
To address the problem, there are studies that further adopt deep learning-based object detectors (e.g., Fast R-CNN \cite{nips-2015-fast-rcnn}) to improve region detection
and then achieve outstanding RRG performance on benchmark datasets.
%
For example, Tanida \textit{et al.} \cite{cvpr-2023-interactive} leverage the Chest ImaGenome dataset \cite{nips-2023-chest-imagenome-dataset} with 29 annotated anatomical regions to fine-tune an anatomy-based object detector that is initialized from ImageNet \cite{deng2009imagenet}, and further use it to provide regional visual features to guide radiology report generation.
To summarize, regional visual features allow RRG models to leverage fine-grained visual signals 
from radiographs
to generate more precise texts so that ensuring them containing details of particular pathology information in the final reports.

\begin{figure*}[t]
\centering
\subfigure[]{
\includegraphics[height=0.33\linewidth]{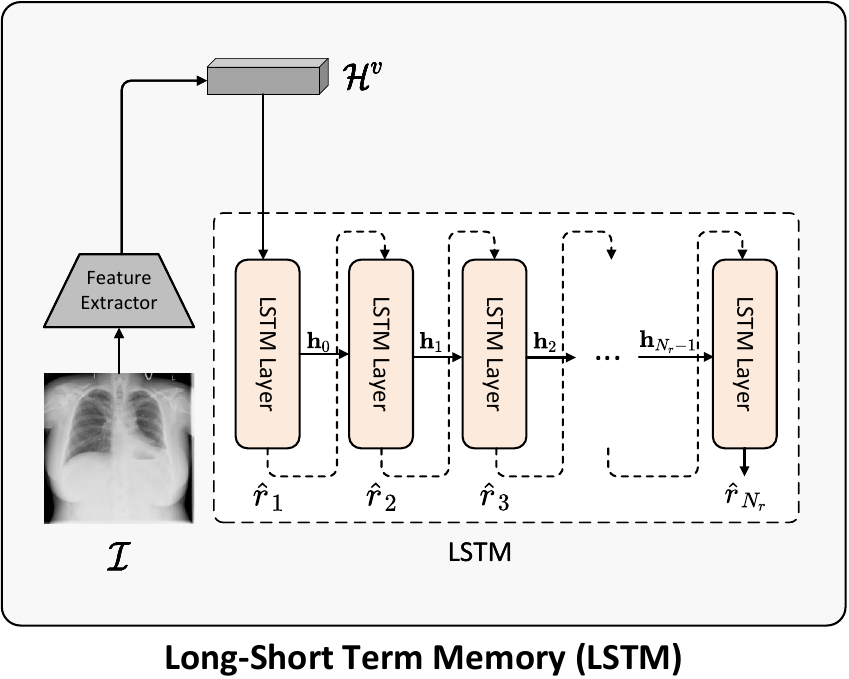}
\label{sfig: lstm}}
\subfigure[]{
\includegraphics[height=0.33\linewidth]{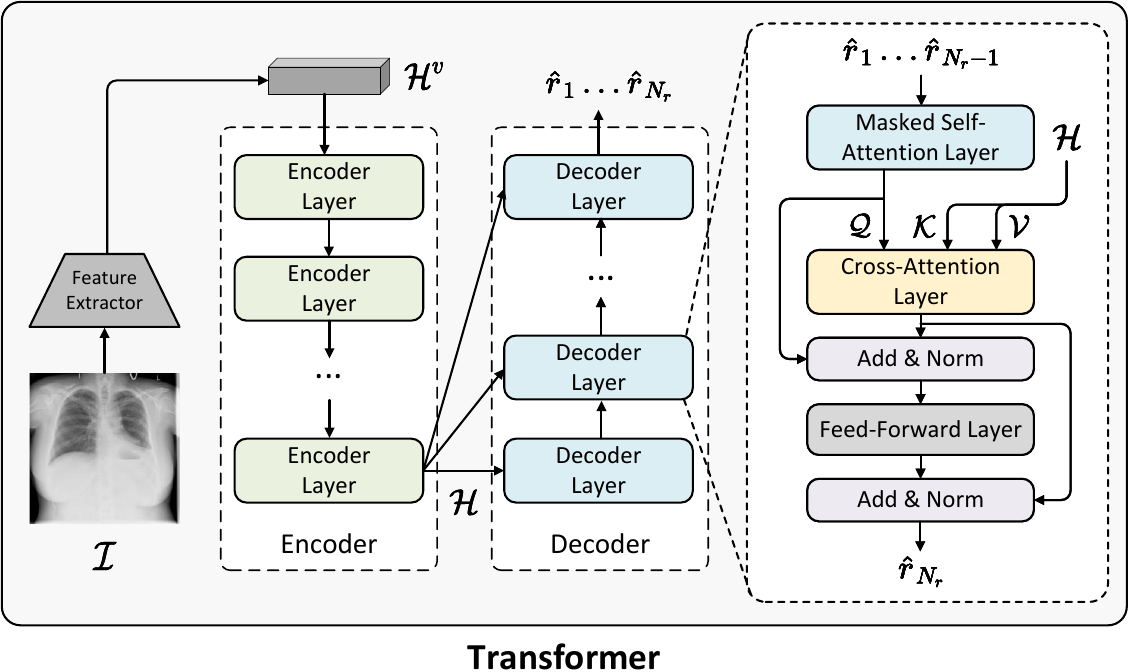}
\label{sfig: trans}}
\vspace{-0.25cm}
\center
\caption{The architecture of autoregressive models for textual-only approaches, including (a) long-short term memory (LSTM) and (b) Transformer.}
\vspace{-0.25cm}
\end{figure*}

\subsubsection{Global-Regional Aggregated Visual Features}

Consider that employing visual features globally or locally has demonstrated its advantages,
%
it is thus natural to further consider join both global and regional visual features to enhance RRG, with the global features ensuring the overall consistency of the generated report, and the regional features facilitating the preciseness of produced texts to describe specific diseases.
However, it is not trivial to combine them since global and regional features differ from each other in both contents and structures, which causes the feature misalignment problem.
%
Towards solving this problem, recent studies \cite{cov-ctr} are developed to integrate global and regional features
%
by adopting a dual-branch feature extraction pipeline to process the input radiograph, and fuse the extracted features from both branches into an integrated one, which is then used by subsequent networks to provide signals from different views (i.e., global and regional) for the report generation process.
%
In other words, the visual feature extractor $f_{fe}$ consists of three components for approaches using global-regional aggregated visual features, namely, a global feature extractor $f^g_{fe}$, a regional feature extractor $f^r_{fe}$, and a fusion process $f_a$ 
to combine the global and regional features to get $\mathcal{H}$, whose process is illustrated in Figure \ref{sfig: global-regional-aggregated-visual-feature}.
%
Specifically, given a radiograph $\mathcal{I}$, the global branch employs $f^g_{fe}$ to extract the global visual feature $\mathcal{H}^g$ from $\mathcal{I}$;
the regional branch utilizes $f^r_{fe}$ to first obtain specific regions from $\mathcal{I}$, and then extract regional visual features $\mathcal{H}^r$ from the resulting regions;
and $f_{a}$ is used to integrate $\mathcal{H}^g$ and $\mathcal{H}^r$ into $\mathcal{H}$.
Thus, the overall process is formulated by
\begin{equation} \label{eq: global_regional}
    \mathcal{H} = f_{a} \left( \mathcal{H}^g, \mathcal{H}^r \right), \ \ \mathcal{H}^g = f^g_{fe} \left( \mathcal{I} \right), \ \  \mathcal{H}^r = f^r_{fe} \left( \mathcal{I} \right)
\end{equation}
where $\mathcal{H}$ is then processed by $f_{ve}$ into $\mathcal{H}^v$, so that $\mathcal{H}^v$ is able to facilitate the report generation process with aggregated information from both global and regional views of $\mathcal{I}$.
%
As a representative study, Li \textit{et al.} \cite{cov-ctr} extract both global and regional features from input radiographs, and integrate them 
to facilitate RRG.
Specifically,
extracting global features in this approach employs a DenseNet-121 (a variant of DenseNet with 121 layers) as $f^g_{fe}$ to extract $\mathcal{H}^g$ from $\mathcal{I}$.
%
For regional features, it selects important regions in the radiograph through a masking mechanism, where features lower than a threshold in the representation of $\mathcal{I}$ are filtered out and the rest features are kept, and
then uses another DenseNet-121 (serving as $f^r_{fe}$) to extract regional visual features $\mathcal{H}^r$.
Finally, an element-wise summation operation $f_a$ is adopted to fuse $\mathcal{H}^g$ and $\mathcal{H}^r$ and obtain the integrated feature $\mathcal{H}$ encoded by $f_{ve}$ to facilitate report generation.
Similarly, Wang \textit{et al.} \cite{Wang2022SelfAG} consider both global and regional anatomical regions of radiographs for RRG, and propose a self-adaptive fusion gate module to dynamically aggregate global and regional visual features, so as to improve RRG. 
In leveraging both global and regional visual features, RRG models are able to have a comprehensive understanding of the overall information and the specific details of the radiograph, and generate better reports compared with that only enhanced by one of the feature types.

\subsection{Textual-only Approaches}

As the target of the RRG task, reports reveal some special characteristics of the texts associated to radiographs.
First, they usually 
contain enriched medical information such as particular terms in describing diseases and organs,
where such medical information are highly correlated with each other and indicate relationships among specific entities (e.g., diseases).
Second, they are normally highly patternized, with significant structural information (e.g., formats or templates for the entire report or descriptions of diseases) to formulate the entire report.
Third, they are usually long comparing to other type of texts such as normal image captions, especially when requiring to describe complicated abnormalities in the radiographs, leading to the error-prone report writing process for physicians.
%
To consider the aforementioned properties of radiology reports, there are textual-only approaches proposed by employing different information to facilitate the report generation process.
Normally, existing textual-only approaches are formulated in the following manner.
Once the input radiograph $\mathcal{I}$ is processed by a visual encoder,
an information extractor $f_{ie}$ is adopted to extract information $\mathcal{K}$ from $\mathcal{H}^v$ and a text generator $f_{tg}$ takes $\mathcal{H}^v$ and $\mathcal{K}$ to produce the final report $\widehat{\mathcal{R}}$ by
\begin{equation} \label{eq: textual-only}
     \widehat{\mathcal{R}} = f_{tg} \left( \mathcal{H}^v, \mathcal{K} \right), \ \ \mathcal{K} = f_{ie} \left( \mathcal{H}^v \right)
\end{equation}
%
%
Existing approaches usually adopt autoregressive models (e.g., LSTM \cite{HochSchm97} and Transformer \cite{vaswani2017attention}) as the $f_t$
to produce texts (i.e., normally words) sequentially.
For example, approaches 
\cite{jing-etal-2018-automatic, jing-etal-2019-show, aaai-2020-when, iccv-2021-visual-textual, cvpr-2021-self-boosting} utilizing LSTM as $f_t$ use the extracted visual feature as the initial hidden state of LSTM, and predict descriptive words one by one for the final report, with the details of their model architecture demonstrated in Figure \ref{sfig: lstm}.
Other approaches \cite{aaai-2019-knowledge, chen-etal-2020-generating, chen-etal-2021-cross-modal, yan-etal-2021-weakly-supervised, miura-etal-2021-improving, nishino-etal-2022-factual, you-etal-2022-jpg, qin-song-2022-reinforced, cov-ctr, cvpr-2023-interactive} use Transformer as $f_t$ to produce the entire report through sequentially predicting the next word at each step based on multi-head attention mechanism, whose details are demonstrated in Figure \ref{sfig: trans}.
According to the aforementioned characteristics of radiology reports and the approaches proposed to address them,
we categorize existing textual-only approaches into three groups, including knowledge enhancement, report structuralization, and progressive generation, whose details are illustrated as follows with elaborated descriptions.

\subsubsection{Knowledge Enhancement}

Radiology reports normally contain crucial medical information, such as medical terms and their relations, which are essential for presenting important facts in the reports.
Specifically, medical terms are critical in identifying specific disease categories and attributes, as manifested in radiographs,
while relations among these terms not only reveal the associations between diseases and affected organs but also provide valuable insights to them, 
they both provide significant enhancement to the quality and accuracy of reports.
%
Therefore, knowledge enhancement-based approaches \cite{aaai-2019-knowledge, aaai-2020-when, cvpr-2021-exploring, you-etal-2022-jpg, nishino-etal-2022-factual, kale-etal-2023-replace, kale-etal-2023-kgvl, cvpr-2023-dynamic, hou-etal-2023-organ, cvpr-2023-dynamic} are motivated and facilitate RRG through discovering and leveraging such crucial information for RRG,
particularly through medical terms, entities and relations, as well as external help of knowledge graphs.
Details of these approaches are introduced
in the following texts.

\smallbreak\noindent\textbf{Medical Terms}
are generally performed as one of the most commonly used knowledge types in radiology reports, which reveal detailed descriptions of particular diseases.
Therefore, for RRG, medical terms are able to provide additional hints for generating descriptions of specific diseases and their corresponding attributes.
Existing approaches mainly train an image tagger to extract medical terms $\{ \widehat{t}_1 \dots \widehat{t}_{N_t} \}$ from radiographs, and use the terms as the information $\mathcal{K} = \{ \widehat{t}_1 \dots \widehat{t}_{N_t} \}$ to enhance RRG.
Therefore, the process for RRG is formulated as
\begin{equation}
    \{ \widehat{t}_1 \dots \widehat{t}_{N_t} \} = f_{ie} \left( \mathcal{H}^v \right)
\end{equation}
and
\begin{equation}
    \widehat{\mathcal{R}} = f_{tg} \left( \mathcal{H}^v, \{ \widehat{t}_1 \dots \widehat{t}_{N_t} \} \right)
\end{equation}
%
where, in order to improve the quality of extracted medical terms, $f_{ie}$ is often trained on auto-annotated data.
In doing so, people adopt existing medical term annotators, e.g., CheXpert \cite{aaai-2019-chexpert} and MeSH\footnote{\url{https://www.nlm.nih.gov/mesh/meshhome.html}}, to extract $N_{t^*}$ medical terms $\{ t^*_1 \dots t^*_{N_{t^*}} \}$ from the gold standard report $\mathcal{R}^*$, and use them as the input to train $f_{ie}$, which allows $f_{ie}$ to extract medical terms that are more likely to appear in the report from the radiograph features and thus provide better guidance for RRG.
For example,
You \textit{et al.} \cite{you-etal-2022-jpg} perform RRG through disease prediction.
They firstly extract visual features from the input radiograph $\mathcal{I}$ and construct a joint semantic space to align both visual and textual features, resulting in integrated global visual representation $\mathcal{H}^v$.
Then, they adopt a classification layer and GRU layers as $f_{ie}$ to predict the medical terms $\{ \widehat{t}_1 \dots \widehat{t}_{N_t} \}$ based on $\mathcal{H}^v$, which results in the representation (denoted as $\mathcal{H}^t$) of $\{ \widehat{t}_1 \dots \widehat{t}_{N_t} \}$.
To optimize $f_{ie}$, this approach minimizes the negative log likelihood between model predictions $\{ \widehat{t}_1 \dots \widehat{t}_{N_t} \}$ and the gold standard medical terms $\{ t^*_1 \dots t^*_{N_{t^*}} \}$.
%
To predict $\{ \widehat{t}_1 \dots \widehat{t}_{N_t} \}$, the approach adopts two types of labels as gold standard labels, namely, MeSH-annotated labels and context-aware labels, which are extracted by an external medical labeler named MeSH and by human annotators from the gold standard report, respectively.
%

%
However, directly predicting disease labels fails to judge the status of the predicted disease categories, which is still limited in facilitating the report generation process.
Therefore, Kale \textit{et al.} \cite{kale-etal-2023-replace} aim to predict medical terms from visual features and perform a disease classification upon the estimated terms.
The approach adopts an image tagger $f_{ie}$ to classify whether the diseases are presented in the radiograph. 
In doing so, it employs the visual feature extractor $f_{fe}$ (i.e., ResNet-50 \cite{he2016residual}) in Eq. (\ref{eq: visual_only_approach}) to extract the visual feature $\mathcal{H}^v$ from the input radiograph $\mathcal{I}$, and utilize several fully-connected projection layers to compute the scores $\{ c_1 \dots c_{N_t} \}$ of different medical terms $\{ t^*_1 \dots t^*_{N_t} \}$ based on $\mathcal{H}^v$.
Herein, each score $c_i \in \{ c_1 \dots c_{N_t} \}$ is converted into binary value (i.e., $c_i$ equals to $0$ or $1$) based on a cutoff threshold $T$, where $c_i=1$ and $c_i=0$ indicate that the disease is presented or absent, respectively.
The corresponding representation $\mathcal{K}$ of the medical terms is then encoded into latent representation to generate the final report $\widehat{\mathcal{R}}$.
To obtain $\{ t^*_1 \dots t^*_{N_t} \}$, the approach fine-tunes a CNN-based classifier (i.e., ResNet) and CheXpert to annotate medical terms.
%
Similarly, Yuan \textit{et al.} \cite{yuan2019automatic} propose to predict medical terms and fuse the resulting features with visual features, so as to enhance RRG.
It is worth noting that the approach considers both frontal and lateral views of the radiographs to perform RRG, where the radiographs are concatenated and used as a single radiograph for further processing.
%
Similarly, Wang \textit{et al.} \cite{miccai-2022-wang-etal-semantic-assisted} adopt predicted medical terms from the extracted visual feature of radiographs, and use the representation of the predicted terms to offer semantic assistance for the report generation process.  
Harzig \textit{et al.} \cite{harzig2019addressing} classify whether the generated sentences are describing an abnormal case, so as to facilitate the model in generating more precise descriptive sentences.
Furthermore, Nishino \textit{et al.} \cite{nishino-etal-2022-factual} perform a planning-based report generation based on annotated medical terms.
The approach first employs a medical term annotator (i.e., VisualCheXBERT \cite{visualchexbert}) as $f_{ie}$ to extract medical terms $\{ \widehat{t}_1 \dots \widehat{t}_{N_t} \}$ from radiographs, and adopts a content planner to process $\{ \widehat{t}_1 \dots \widehat{t}_{N_t} \}$ into a series of plans, where each plan suggests the order in which the terms appear.
And the approach utilizes a text generator (i.e., T5 \cite{2020t5}) to produce descriptive sentences according to the plans, and construct the final report according to the plan order.
Then, it adopts a content planner to process $\{ \widehat{t}_1 \dots \widehat{t}_{N_t} \}$ into a series of plans,
which suggest the order in which the terms appear.
Finally, the approach utilizes a text generator to produce descriptive sentences based on the plans, and concatenates all sentences to get the final report.

\smallbreak\noindent\textbf{Entities and Relations}
are pivotal elements to characterize the associations between different components in text.
In the setting of RRG, entities normally denote a series of medical terms (such as disease categories, organs, attributes, etc.), and relations usually indicate the associations among different terms (such as anatomical location, diagnostic indication, disease attributes, health status, etc.),
%
which offer direct assistance for the report generation process.
To leverage the information, existing approaches \cite{dalla-serra-etal-2022-multimodal} formulate entities and relations by a list of tuples $\mathcal{U} = u_1 \cdots u_{N_e}$ where the $n_e$-th tuple $u_{n_e} = (e_{n_e, 1}, r_{n_e}, e_{n_e, 2})$ with $e_{n_e, 1}$, $e_{n_e, 2}$ denoting entities, and $r_{n_e}$ referring to their relation.
Herein, $\mathcal{U}$ is utilized as extra information $\mathcal{K}$ to facilitate the report generation process along with entities and relations.
%
For example, Dalla Serra \textit{et al.} \cite{dalla-serra-etal-2022-multimodal} 
%
firstly utilize RadGraph \cite{radgraph} and ScispaCy \cite{neumann-etal-2019-scispacy} to extract entities and relations from the gold standard report $\mathcal{R}^*$.
%
Herein, the entities represent the abnormalities and organs, and the relations indicate their relationships, including ``\textit{suggestive of}'', ``\textit{located at}'', and ``\textit{modify}''.
Then, the approach employs a tuple extractor $f_{te}$ to predict the entity and relation tuples based on the extracted feature $\mathcal{H}^v$ from the input radiograph $\mathcal{I}$.
Afterwards, the tuple is used as the condition of report generation, where the text decoder $f_{t}$ generates the final report $\widehat{\mathcal{R}}$ based on $\mathcal{H}^v$ and $u_{n_e}$.

\smallbreak\noindent\textbf{Knowledge Graph}
records structural relationships among different term-based information, which usually consists of nodes and edges 
that directly indicate the internal correlation between diseases and organs in the medical field,
thus is potential to help RRG with effective guidance.
However, constructing a decent knowledge graph for RRG is not a trivial task, where the settings of its nodes and edges should be reasonable and helpful to the RRG process.
Existing approaches to building the knowledge graph are categorized into two groups, namely, prediction-based approaches and pre-construction-based approaches.
The former approaches usually predict the probability value of nodes and edges based on extracted visual features;
%
the latter ones pre-define the nodes and edges of knowledge graphs based on clinical rules.
They usually construct the nodes to represent specific disease categories or organs, and construct the edges to refer to the correlations among diseases or associations among diseases and particular organs.
Once the knowledge graph $\mathcal{G}$ is constructed, the aforementioned approaches employ a graph encoder $f_{ie}$ to extract information $\mathcal{K}$ from $\mathcal{G}$ with the guidance of $\mathcal{H}^v$, and then utilize the text generator $f_{tg}$ to generate the final report.
In performing the aforementioned processes, Li \textit{et al.} \cite{aaai-2019-knowledge} firstly propose to perform a knowledge-enhanced RRG through three main steps, namely, encoding, retrieving, and paraphrasing.
Given a radiology image $\mathcal{I}$, the encoding step of the approach extracts the visual feature $\mathcal{H}^v$ from $\mathcal{I}$, and encodes $\mathcal{H}^v$ into an abnormality graph $\widehat{\mathcal{G}}$.
The node and edges in $\widehat{\mathcal{G}}$ are compared with the gold standard graph $\mathcal{G}^*$, where the binary cross-entropy loss is computed and used to optimize the model.
To obtain $\mathcal{G}^*$, the approach defines clinical abnormalities stemming from thoracic organs as nodes of $\mathcal{G}^*$, and uses the co-occurrence of abnormalities as edges.
Then, the retrieving step obtains report templates based on the predicted knowledge graph, and the paraphrasing step refines the retrieved templates with enriched descriptions so as to generate the final report $\widehat{\mathcal{R}}$ accordingly.

However, approaches based on predicted knowledge graphs present a limitation where they are only able to predict a part of the entire graph, which makes it hard for them to get fully enhanced by knowledge graphs to improve RRG.
To address the limitation, there are studies that pre-construct the knowledge graphs instead of predicting them, so as to provide high-quality knowledge graphs to assist the report generation process.
In doing so, Zhang \textit{et al.} \cite{aaai-2020-when} manage to utilize a pre-constructed knowledge graph to assist the report generation process.
The approach firstly pre-trains an image classification model $f_{fe}$ on CheXpert \cite{aaai-2019-chexpert} to classify a series of medical findings corresponding to the nodes of a pre-defined knowledge graph $\mathcal{G}$, where $f_{fe}$ is used to extract the visual feature $\mathcal{H}^v$ from the input radiograph $\mathcal{I}$.
Then, it conducts the attention mechanism over $\mathcal{H}^v$ to initialize the knowledge graph $\mathcal{G}$, and employs graph convolutional networks (GCN) \cite{kipf2017semisupervised} to propagate information across nodes and edges of $\mathcal{G}$, where $\mathcal{G}$ is further optimized through a multi-label classification process.
Finally, $\mathcal{G}$ is encoded into graph embedding $\mathcal{H}^G$ for the text decoder, which generates the final report $\widehat{\mathcal{R}}$ with the assistance of $\mathcal{H}^G$. 
To define $\mathcal{G}$, the approach extracts 20 categories of keywords from the radiology reports as the nodes of $\mathcal{G}$, and utilizes its edges to represent the relationship between diseases and particular organs.
The knowledge graph constructed by Zhang \textit{et al.} is widely exploited in the follow-up studies \cite{cvpr-2021-exploring, cvpr-2023-dynamic}.
Specifically, Liu \textit{et al.} \cite{cvpr-2021-exploring} further utilize medical terms as prior knowledge, and adopt related reports and pre-constructed knowledge graph as posterior knowledge to enhance RRG.
Kale \textit{et al.} \cite{kale-etal-2023-kgvl} construct a knowledge graph for X-ray images to augment a pre-trained visual-language BART \cite{lewis-etal-2020-bart}, so as to improve RRG.
%
%
%
Particularly, the approach constructs $\mathcal{G}$ based on extracted term and relation tuples from radiology reports following the approach of Kale \textit{et al.} \cite{kale2022knowledge}.
It defines eight logical relations based on a series of medical characteristics, including anatomy, entities type, anatomical locations and properties, findings, etc., as the edges of $\mathcal{G}$, and integrates several categories based on Radlex entities\footnote{\url{http://radlex.org/}} as the nodes of $\mathcal{G}$.
%
%
Recent studies consider the deficiency of pre-constructed knowledge graphs in conventional graph-based approaches, where the pre-defined scope of knowledge hurts the effectiveness of these approaches.
Therefore, Li \textit{et al.} \cite{cvpr-2023-dynamic} propose to extend the pre-constructed knowledge graph, so as to facilitate the RRG process with both pre-defined and additional knowledge.
The approach regards the pre-constructed knowledge graph $\mathcal{G}$ \cite{aaai-2020-when} as general knowledge, and performs a dynamic graph construction process upon $\mathcal{G}$ to introduce report-specific knowledge into $\mathcal{G}$. 
In doing so, the approach firstly retrieves top-$K$ the most similar reports based on the similarity between the extracted visual feature $\mathcal{H}^v$ and report feature $\mathcal{H}^r$.
Then, it extracts anatomy and observation entities by Stanza \cite{qi-etal-2020-stanza}, and utilizes the resulting entities to obtain report-specific knowledge $\mathcal{K}$ through RadGraph \cite{radgraph}.
To extend the pre-constructed knowledge graph $\mathcal{G}$, the approach employs relations in RadGraph as new edges of the extended graph $\widehat{\mathcal{G}}$, and adds $\mathcal{K}$ as additional nodes of $\widehat{\mathcal{G}}$.
Finally, $\widehat{\mathcal{G}}$ is capable of facilitating the report generation process with both general and report-specific knowledge, resulting in the generated report $\widehat{\mathcal{R}}$.
Similarly, Yan \textit{et al.} \cite{yan2023attributed} enrich the constructed knowledge graph with abnormality nodes and attribute nodes, and propose an automatic approach to build the knowledge graph based on annotations, radiology reports, and RadLex radiology lexicon.

In leveraging different types of information, RRG models are capable of generating better reports with more focus on particular information, e.g., specific terms and relationships between terms.
Among this information, medical terms provide term-level information for report generation, entities and relations further indicate the relationship between every two terms, and the knowledge graph offers a holistic view of different terms (nodes) and their relationships (edges), where more information are provided for the report generation process from medical terms to knowledge graph, thereby generating reports of higher elaboration.

\subsubsection{Report Structuralization}

One of the most remarkable characteristics of radiology reports is that reports are usually highly patternized.
Specifically, physicians normally write descriptions following particular sentence structures of diseases and organs, and follow certain templates to construct the entire report.
This observation indicates that radiology reports consist of significant structure information, which presents the potential to enhance the report generation process by filling in enriched details in describing diseases based on the confirmed structures.
Therefore, some studies \cite{aaai-2019-knowledge, yan-etal-2021-weakly-supervised, dalla-serra-etal-2022-multimodal, kale-etal-2023-replace} 
propose to process radiology reports into particular patterns based on pre-defined rules, where the report generation process is then guided by these patterns so as to enrich the description details accordingly.
Particularly, these studies adopt two main categories of patterns to facilitate RRG, namely, report template and report clustering.
Details of each specific category are illustrated in the following texts.

\smallbreak\noindent\textbf{Report Templates}
provide guidance for RRG that is neglected by entities and relations.
%
%
Normally, the approach based on report template firstly determines the overall structure of the generated report and then allows RRG approaches to fill in descriptive contents into these templates according to the radiology image.
In doing so, existing approaches \cite{aaai-2019-knowledge, kale-etal-2023-replace} mainly employ descriptive sentences of the same abnormality as report templates for RRG.
They perform RRG in a two-stage manner, including template construction and report generation processes.
The template construction process obtains the report template by extracting patterns from all radiology reports in the entire training set, and the report generation process fills descriptive details into the report template, thereby producing the final report.

Specifically, Li \textit{et al.} \cite{aaai-2019-knowledge} utilize similar descriptive sentences of different abnormalities as templates to assist the report generation process.
For each specific abnormality, the approach collects all sentences that describe the abnormality in the training corpus, where the sentences with the same meaning are grouped, and the most frequent one is used as the template.
Given the input radiograph $\mathcal{I}$, the report template is able to serve as guidance for generating descriptions for each particular abnormality, thereby constructing the entire report $\widehat{\mathcal{R}}$ according to $\mathcal{I}$.
Later, Kale \textit{et al.} \cite{kale-etal-2023-replace} construct a more fine-grained report template, with specific labels of described organs as templates for the contents of sentences in the final report.
For example, each template is represented by a particular term, indicating the abnormality region that the sentence is describing, e.g., ``\textit{lung1}'' refers to the first template type to describe the lung area.
From another aspect, while existing studies adopt pre-defined report templates for RRG, Chen \textit{et al.} \cite{chen-etal-2020-generating} propose relational memory networks to automatically store the report template information and use it to guide the report generation process.
%
Particularly, the relational memory networks are conducted as a matrix $\mathcal{M}_t$, where $\mathcal{M}_t$ at the $t$-th generation step is updated to $\mathcal{M}_{t+1}$ for the next step during the generation processes of reports.
Also, the approach adopts a memory-conditional layer normalization mechanism to retrieve the template information from the memory matrix and incorporate it into the report generation process, so as to facilitate RRG with the recorded templates.
Similarly, Yu \textit{et al.} \cite{icbb-2022-yu-etal-clinically-coherent} adopt a memory bank to preserve key information for radiology report generation, and employ reinforcement learning to improve the model to generate clinically accurate reports that contain critical information, such as the presence status of different diseases.

\smallbreak\noindent\textbf{Report Clustering}
considers features shared by similar reports (e.g., relevant phrases or identical expressions) that are associated with the same or similar radiographs, which provides a reference for physicians when writing reports even if an input radiograph is previously unseen.
In doing so, the existing approach clusters all radiology reports in the training corpus into a series of categories, where reports in the same category are considered to be semantically close to each other.
Thus, the report clustering process provides additional supervision signals for the optimization of the RRG pipeline, thereby facilitating the report generation process.

Specifically, Yan \textit{et al.} \cite{yan-etal-2021-weakly-supervised} propose a weakly supervised contrastive learning with clustered reports to facilitate RRG.
In doing so, the approach first projects all radiology reports in the training corpus onto a semantic space through CheXBERT \cite{smit-etal-2020-combining}, and conducts a K-means clustering algorithm upon the report representations.
Considering that reports in the same category are semantically similar to each other, the approach optimizes the model to generate reports that are close to the semantically close ones through contrastive learning.
Then, the similarity of positive pairs is maximized and the one of negative pairs is minimized in a way of contrastive learning, which finally assists the overall pipeline to generate an image-text aligned report $\widehat{\mathcal{R}}$.

In performing report structuralization for RRG, RRG models are able to utilize the pattern information shared by different reports to assist the report generation process.
Specifically, report templates are carefully constructed according to the radiology reports, which requires an additional manual selection process to improve RRG models.
Report clustering acts as an automatic solution to extract pattern information from reports and provides heuristics upon automatic report structuralization.

\begin{figure*}[t]
\centering
\subfigure[]{
\includegraphics[height=0.1325\linewidth]{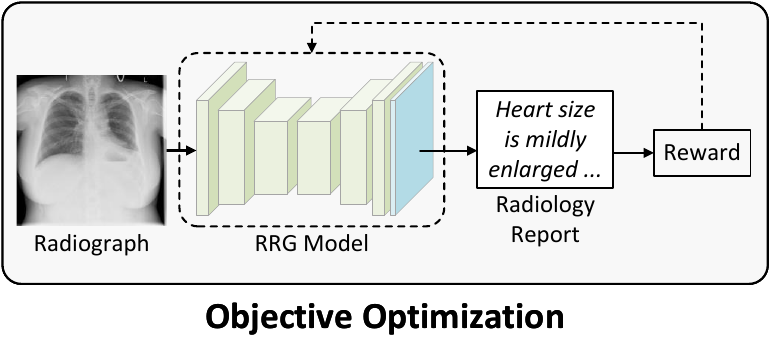}
\label{sfig: objective-optim}}
\subfigure[]{
\includegraphics[height=0.1325\linewidth]{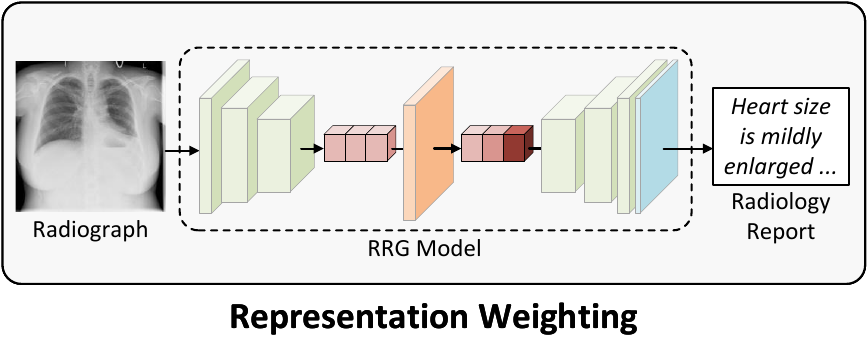}
\label{sfig: representation-weighting}}
\subfigure[]{
\includegraphics[height=0.1325\linewidth]{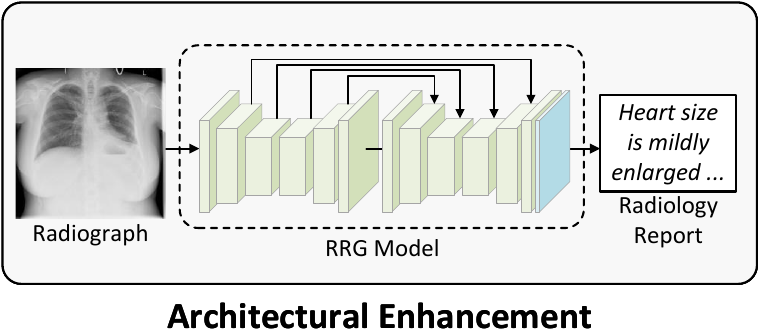}
\label{sfig: architecture-enhancement}}
\vspace{-0.20cm}
\center
\caption{The architectures of three main categories of cross-modal approaches that enhance the cross-modal alignment for RRG from 
different perspectives, including (a) objective optimization, (b) representation weighting, and (c) architecture enhancement.}
\label{sfig: cross-modal}

\vspace{-0.25cm}
\end{figure*}

\subsubsection{Progressive Report Generation}

Although existing textual-only approaches have adopted a series of textual characteristics for RRG, the one-step generation process of the entire report is still challenging.
Therefore, progressive report generation is promoted, which decomposes the RRG task into multiple sub-tasks, so as to alleviate the difficulty of generating full radiology reports.
In doing so, existing approaches \cite{xue2018multimodal, nooralahzadeh-etal-2021-progressive-transformer} manage to perform RRG in a two-stage manner.
Given a radiology image $\mathcal{I}$, they firstly generate an intermediate report $\widehat{\mathcal{Z}}$ (e.g., high-level summarization of the entire report) according to $\mathcal{I}$, and produce the final report $\widehat{\mathcal{R}}$ according to $\widehat{\mathcal{Z}}$ progressively.

Specifically, Xue \textit{et al.} \cite{xue2018multimodal} manage to generate radiology reports in a sentence-by-sentence paradigm, where the study adopts the previously produced sentence as the condition to generate the next one, thereby ensuring the textual coherence in the resulted report.
Compared to sentence-level progressive generation, Nooralahzadeh \textit{et al.} \cite{nooralahzadeh-etal-2021-progressive-transformer} propose a consecutive generation framework to perform a two-stage report generation process.
The approach firstly processes the gold standard report $\mathcal{R}^*$ to sequentially extract the disease words, negation or uncertainty of each word, and disease attributes from $\mathcal{R}^*$ based on dependency graph parsing, which results in summarization with respect to disease categories and attributes
Given the input radiograph $\mathcal{I}$, the approach utilizes a language model (i.e., BART \cite{lewis-etal-2020-bart}) to 
firstly produce summarization, and then generates the final report $\widehat{\mathcal{R}}$ according to the summarization.
By decomposing the entire report generation process into multiple steps, RRG models are able to perform RRG in a coarse-to-fine manner, thereby generating more precise report than the conventional one-step generation paradigm.

\subsection{Cross-modal Approaches}

Owing to the cross-modal nature of RRG, it is straight-forward to consider leveraging the relationship between the two modalities, i.e., vision (radiographs) and text (reports),
where
such relationship actually establishes the connection between the characteristics from both sides.
Yet, utilizing the radiograph-report connection is not a trivial task, where the cross-modal mapping has its own characteristics in the following aspects.
First, human preferences play a pivotal role in choosing different ways to employ such connection, where physicians usually
construct reports based on their own knowledge such as clinical experiences.
Second, radiograph regions usually receive imbalanced attentions in report writing, so that leads to different descriptive sentences with variant levels of details in the final report.
Therefore
to consider these characteristics, cross-modal approaches are motivated to optimize or enhance the alignment between radiographs and reports to facilitate RRG.
Specifically, some approaches \cite{nips-2018-hrgr-agent, nishino-etal-2020-reinforcement, cvpr-2021-self-boosting, delbrouck-etal-2022-improving, liu-etal-2021-competence} aim to improve the optimization objectives of the RRG process from both modalities
by requiring additional supervision signals (e.g., medical term annotations).
Meanwhile, other studies \cite{jing-etal-2018-automatic, chen-etal-2020-generating, iccv-2021-visual-textual, liu-etal-2021-contrastive, chen-etal-2021-cross-modal, qin-song-2022-reinforced} conduct representation weighting for different modalities in order to balance the contribution from either visual or textual guidance, thus enhance fine-grained alignment between radiographs and reports to facilitate RRG.
Later,
based on the aforementioned work, recent approaches \cite{cvpr-2023-kiut, cvpr-2023-metransformer}
enhance the cross-modal encoding and decoding from the perspective of model architecture,
and integrate additional information to further improve the encoder and decoder for RRG.
Therefore, we categorize cross-modal approaches into three groups, emphasizing on objective optimization, representation weighting, and architecture enhancement. 
These approaches in three categories are illustrated in the following texts with detailed information.

\subsubsection{Objective Optimization}
Objective optimization-based approaches \cite{jing-etal-2018-automatic, jing-etal-2019-show, chen-etal-2020-generating, nishino-etal-2020-reinforcement, chen-etal-2021-cross-modal, cvpr-2021-self-boosting, liu-etal-2021-competence, iccv-2021-visual-textual, yan-etal-2021-weakly-supervised, liu-etal-2021-contrastive, delbrouck-etal-2022-improving, qin-song-2022-reinforced, cvpr-2023-kiut, cvpr-2023-metransformer} enhance the cross-modal alignment between radiographs and reports by improving the training objectives, so as to facilitate RRG.
The architecture of this approach is shown in Figure \ref{sfig: objective-optim}.
We classify existing approaches into four main categories to optimize the training objectives, including reinforcement learning, curriculum learning, self-boosting, and pre-training whose details are illustrated in the following texts.

\smallbreak\noindent\textbf{Reinforcement Learning}
provides an alternative learning objective rather than
the standard supervised learning that typically employs a cross-entropy loss function for optimization (as shown in Eq. (\ref{eq: cross-entropy})), which is prone to exposure bias \cite{sequence-level-training} and does not always align with evaluation metrics \cite{sequence-level-training}.
In contrast, reinforcement learning is able to directly optimize performance based on different metrics by using these metrics as rewards to update model parameters. 
The key elements of reinforcement learning include an agent, environment, action, policy, and reward. 
Several approaches for RRG \cite{nips-2018-hrgr-agent, jing-etal-2019-show, nishino-etal-2020-reinforcement, delbrouck-etal-2022-improving} treat the generation model as the agent interacting with an external environment (i.e., features in different modalities), where the model parameters $\theta$, which is represented as policy $p_\theta$, dictate the actions taken, such as predicting the entire report or the next token.
These approaches focus on designing various evaluation metrics, such as BLEU \cite{papineni-etal-2002-bleu}, METEOR \cite{banerjee-lavie-2005-meteor}, and ROUGE \cite{lin-2004-rouge}, and use the combination of them as rewards $r$ to optimize the generation model.
The primary objective of reinforcement learning is to maximize the expected reward for a prediction $\widehat{y}$ made by the model, which is expressed as
\begin{equation} \label{eq: reinforce-reward}
L(\theta) = \mathbb{E}_{\widehat{y}\sim p_\theta} \left[ r(\widehat{y}) \right]
\end{equation}
The gradient of $L(\theta)$ with respect to $\theta$ is computed using the REINFORCE algorithm \cite{williams-1992-reinforce} through
\begin{equation} \label{eq: reinforce-gradient}
\nabla_\theta L(\theta) = -\mathbb{E}_{\widehat{y}\sim p_\theta} \left[ r(\widehat{y})\nabla_\theta \log p_\theta (\widehat{y}) \right]
\end{equation}
This method ensures that actions resulting in a higher reward $r(\widehat{y})$ are more likely to be repeated in the learning process, guiding the model towards optimal performance.

Li \textit{et al.} \cite{nips-2018-hrgr-agent} adopt a combination of retrieval policy module and generation policy module to optimize RRG, and design a sentence-level and word-level reward for optimization.
Given the input radiograph $\mathcal{I}$, it firstly employs a CNN-LSTM architecture to generate the hidden topic states $\mathcal{H}^r_t$ at $t$ step.
Then, a one-layer perceptron predicts the probability distribution over actions of generation or retrieval based on $\mathcal{H}^r_t$, where the module with higher probability is activated to generate the final report $\widehat{\mathcal{R}}$ or retrieve $\widehat{\mathcal{R}}$ from collected template databases.
Herein, the approach sets a threshold to filter out infrequent sentences, and obtains the template databases according to the frequency of sentences in radiology reports.
Besides, the approach designs a sentence-level reward $r^s$ and word-level reward $r^w$ based on CIDEr \cite{cvpr-2015-cider-metric} following Eq. (\ref{eq: reinforce-reward}) and Eq. (\ref{eq: reinforce-gradient}), so as to optimize the retrieval and generation policy modules, respectively.
Herein, $r^s$ and $r^w$ are computed through delta CIDEr scores upon the entire reports and generated words, respectively.
Jing \textit{et al.} \cite{jing-etal-2019-show} further refine the rewards of Li \textit{et al.} \cite{nips-2018-hrgr-agent}, and proposes rewards of normal and abnormal sentences in a generated report to facilitate the RRG process.
Given the input radiograph $\mathcal{I}$, the approach firstly generates a report $\widehat{\mathcal{R}}$ following the standard RRG process.
Then, it decomposes $\widehat{\mathcal{R}}$ into a series of abnormal and normal sentences, denoting as $S^{ab} = \{ s^{ab}_1 \dots s^{ab}_{N_ab} \}$ and $S^{nr}=\{ s^{nr}_1 \dots s^{nr}_{N_nr} \}$ with $N_{ab}$ and $N_{nr}$ referring to the number of abnormal and normal sentences, respectively.
The rewards $r^{ab}$ and $r^{nr}$ to generate abnormal and normal sentences in $\widehat{\mathcal{R}}$ are computed according to BLEU-4 score \cite{papineni-etal-2002-bleu} through comparing $S^{ab}$ and $S^{nr}$ with the corresponding sentences $S^{*ab}$ and $S^{*nr}$ in the gold standard radiology report $\mathcal{R}^*$, respectively.

Previous reinforcement learning-based studies mainly consider report-level enhancement, where such rewards have limited effectiveness in describing specific diseases.
%
%
Therefore, Nishino \textit{et al.} \cite{nishino-etal-2020-reinforcement} propose to improve the generated reports with a token-level reward, which adopts a clinical reconstruction score (CRS) and a data augmentation approach for reinforcement learning to address the data imbalance difficulties in RRG.
In doing so, the approach adopts a medical labels reconstructor (i.e., BERT \cite{devlin-etal-2019-bert}) following the standard generation process of the final report $\widehat{\mathcal{R}}$, so as to reconstruct the finding labels from $\widehat{\mathcal{R}}$.
Then, CRS is computed as the F1 score between the generated and the input finding labels.
Besides, the approach employs an additional ROUGE-L score \cite{lin-2004-rouge} to design the reward $r$ in Eq. (\ref{eq: reinforce-reward}), where $r$ is written as $r(\widehat{\mathcal{R}}) = \lambda f_{rouge} (\widehat{\mathcal{R}},\mathcal{R}^*) + (1 - \lambda) f_{crs} (\widehat{\mathcal{R}})$ with $f_{rouge} (\cdot)$ and $f_{crs} (\cdot)$ referring to the processes of computing ROUGE-L and CRS scores, respectively.

\smallbreak\noindent\textbf{Curriculum Learning}
\cite{Bengio2009CurriculumL} enables neural networks to gradually proceed from easy samples to more complex ones in training, which is also adopted by the recent study \cite{liu-etal-2021-competence} to gradually enhance the radiograph-report alignment for RRG.
These approaches normally require a well-designed metric to measure the difficulty of the currently underlying task and adjust the training samples accordingly.
However, it is challenging for the RRG task to directly apply the standard curriculum learning paradigm, which measures task difficulties solely relying on a single modality, whereas RRG is a multiple task where both radiographs and reports play essential roles in identifying the difficulty.
Therefore, dedicated curriculum learning is expected for RRG.

To address this challenge, Liu \textit{et al.} \cite{liu-etal-2021-competence} propose a competence-based multi-modal curriculum learning approach with multiple difficulty metrics for RRG.
The proposed approach measures the difficulty $d$ of instances based on their difficulties in capturing and describing the abnormalities from radiographs and radiology reports, which consist of visual and textual difficulty, respectively.
For visual difficulty, the approach considers a radiograph as a more difficult training instance if it contains more abnormalities. 
Specifically, the approach adopts a fine-tuned vision backbone model (i.e., ResNet \cite{he2016residual}) on CheXpert dataset \cite{aaai-2019-chexpert}, which extracts features from all normal radiographs in the training set and computes their average feature $\mathcal{H}^{v,n}$.
Given an input radiograph $\mathcal{I}$, the visual difficulty $d^v$ is defined as the average cosine similarity between the extracted feature $\mathcal{H}^v$ of $\mathcal{I}$ and $\mathcal{H}^{v,n}$, where a higher similarity indicates $\mathcal{I}$ is an easier instance.
For the textual difficulty, the approach adopts the number of abnormal sentences in a report to define its difficulty $d^t$.
Following Jing \textit{et al.} \cite{jing-etal-2018-automatic}, sentences without specific words (e.g. ``\textit{no}'', ``\textit{normal}'', ``\textit{clear}'', and ``\textit{stable}'') are considered to be abnormal sentences, while the others represent normal ones.
For model confidence, the approach also introduces two model-based metrics from perspectives of radiographs and reports, denoting as $d^v_m$ and $d^t_m$, respectively, where $d^v_m$ refers to the entropy value of the predicted probability distribution by the vision backbone model, and $d^v_m$ denotes the negative log-likelihood loss values following \cite{Liu2019ExploringAD, CL-2}.
To optimize the model, the approach builds curriculum learning process based on Plananios \textit{et al.} \cite{liu-etal-2021-competence}, which adjusts the applied $d$ based on perplexity (PPL).

\smallbreak\noindent\textbf{Self-boosting} 
 aims to boost the optimization of different components in the overall pipeline.
In doing so, the overall pipeline is usually conducted in multiple branches with different training objectives, e.g., radiograph-report matching and report generation, which are used to optimize the entire RRG pipeline in an alternative manner.
Herein, when a specific branch is being optimized, the model parameters of other branches are fixed.
Then, the updated branch is enhanced by other branches, which eventually leads to better optimization of the overall RRG pipeline.

Specifically, Wang \textit{et al.} \cite{cvpr-2021-self-boosting} propose a dual-branch framework to enhance RRG in a self-boosting manner.
In doing so, the overall framework of the approach is based on two branches, i.e., a report generation branch and an auxiliary image-text matching branch, where both branches facilitate each other with their internal interactions.
The report generation branch supplies hard negative samples for the image-text matching branch so as to force the latter to perform better feature learning.
The image-text matching branch provides visual-textual aligned features for the report generation branch, which computes a triplet loss based on the cosine similarity of radiograph-report pairs and hard negative samples following the standard process of VSE++ \cite{VSE++}.
In addition, both branches are optimized alternatively in a way that is similar to the training paradigm of generative adversarial networks (GAN) \cite{generative-adversarial-networks}, where the report generation branch focuses on producing the final radiology reports and the image-text matching branch aims to differentiate the generated and gold standard reports.

\smallbreak\noindent\textbf{Pre-training}
is another optimization technique that widely used in representation learning.
Since
most existing RRG approaches tend to train their models from scratch, they often have difficulties in fully optimizing their models, especially the ones with complicated network architecture.
Therefore, some studies adopt pre-trained model weights to provide better initialization for RRG.
For example, Nicolson \textit{et al.} \cite{warm-starting} leverage off-the-shelf pre-trained model weights on the medical domain to enhance radiograph encoding and report generation, where they adopt pre-trained vision transformer (ViT) \cite{dosovitskiy2021an} on ImageNet \cite{deng2009imagenet} and PubMedBERT \cite{pubmedbert} to initialize the visual encoder and text decoder, respectively, so as to transfer the learned knowledge in pre-training to RRG.


In improving the optimization objectives, RRG models are capable of producing more elaborated reports with improved optimization.
Specifically, reinforcement learning and curriculum learning obtain improved performance, which requires carefully designed reward or difficulty metrics for such improvement.
Otherwise, self-boosting serves as a novel approach, which is able to achieve better optimization with dedicated model design.
Although studies that adopt pre-training for RRG are limited, these approaches provide enlightening ideas to perform RRG better.

\subsubsection{Representation Weighting}
Representation weighting-based approaches \cite{jing-etal-2018-automatic, chen-etal-2020-generating, liu-etal-2021-contrastive, cvpr-2021-exploring, iccv-2021-visual-textual, chen-etal-2021-cross-modal, qin-song-2022-reinforced, cvpr-2023-kiut, cvpr-2023-metransformer} aim to enhance the cross-modal alignment through weighting the representations of different modalities for RRG, where the architecture of the approaches is illustrated in Figure \ref{sfig: representation-weighting}. 
Specifically, during the generation process, these approaches compute the visual representation $\mathcal{H}^{v}$ of the radiograph $\mathcal{I}$ and textual representation $\mathcal{H}^{t}$ of the generated reports, and
then calculate the weights for multi-modal representations, which are applied to $\mathcal{H}^{v}$ and $\mathcal{H}^{t}$ so that information from different modalities is aligned by the weighted sum of visual and textual representations.
This process allows RRG models to highlight important information in different modalities for text generation and leverage it accordingly to produce accurate reports.
Generally, representation weighting approaches are categorized into two groups: the first uses attention mechanisms, and the second utilizes memory networks.
Details of the two groups of approaches are illustrated as follows.

\smallbreak\noindent\textbf{Attention Mechanism} \cite{feed-forward-attenion,vaswani2017attention}
serves as a powerful representation weighting approach, which is firstly introduced for machine translation.
Owing to its effectiveness in identifying important contexts in the source language and leveraging them to build a bridge of translation between the source and target languages, the attention mechanism is generalized to many generation tasks, including RRG, to translate source data (e.g., radiology) into the target one (e.g., report) \cite{vinyals2015tell, jing-etal-2018-automatic}.
Details of existing attention-based approaches for RRG are illustrated in the following texts.

Jing \textit{et al.} \cite{jing-etal-2018-automatic} propose a co-attention mechanism for RRG to weight visual representation of radiographs with high-level semantics from medical terms.
In doing so, the approach first employs a visual feature extractor (i.e., VGG-19 \cite{iclr-2015-vgg-net}) to obtain the visual representation $\mathcal{H}^v$ from the input radiograph $\mathcal{I}$, and computes the semantic representation $\mathcal{H}^s$ of medical terms based on $\mathcal{H}^v$.
Then, it adopts the co-attention mechanism to incorporate $\mathcal{H}^v$ and $\mathcal{H}^s$ by applying attention weights to them, where the attention weights are modeled by one-layer perceptrons.
Finally, the weighted representation is sent into a hierarchical LSTM to produce the generated report $\widehat{\mathcal{R}}$.
Furthermore, some studies introduce additional information (e.g., clinical information) into the report generation process through an improved attention mechanism.
In doing so, Liu \textit{et al.} \cite{liu-etal-2021-competence} propose multi-modality semantic attention to establish a correlation between regional image features, semantic representations of key findings, and clinical information.
The approach first extracts the regional visual feature $\mathcal{H}^v$ from the input radiograph $\mathcal{I}$, and obtains the clinical feature $\mathcal{F}^c$ based on the disease, age, and sex of the patient.
Then, it adopts the multi-modality semantic attention to weight the hidden state $\mathcal{H}^t$ at step $t$ by applying the corresponding attention weights of $\mathcal{H}^v$ and $\mathcal{F}^c$ upon them.
Therefore, both regional image feature and clinical information are incorporated into the text decoding process of LSTM, which is able to generate the final report $\widehat{\mathcal{R}}$ that is aligned with $\mathcal{H}^v$ and $\mathcal{F}^c$.

Recent approaches \cite{cvpr-2021-exploring, liu-etal-2021-contrastive} improve the multi-head attention mechanism in Transformer \cite{vaswani2017attention} to facilitate RRG.
Liu \textit{et al.} \cite{liu-etal-2021-contrastive} propose an aggregate attention mechanism to incorporate normal radiographs for RRG, and adopts a contrastive attention mechanism to distinguish abnormalities of radiographs from normal ones, where both attention mechanisms are built based on multi-head self-attention.
In doing so, the approach firstly collects a normality pool $\mathcal{P}=\{ \mathcal{H}^v_1 \dots \mathcal{H}^v_{N_p} \}$ with 1,000 visual features of normal radiographs from the training set.
Given the input radiograph $\mathcal{I}$, it computes the similarity between the extracted representation $\mathcal{H}^v$ of $\mathcal{I}$ and all elements in the normality pool, where weights of the most similar instance are applied upon $\mathcal{H}^v$ resulting in $\mathcal{H}^{v,n}$.
Then, the representation $\mathcal{H}^{v,a}$ of abnormal radiograph is computed through subtracting the representation $\mathcal{H}^v$ of normal image, denoting as $\mathcal{H}^{v,a}=\Vert \mathcal{H}^{v} - \mathcal{H}^{v,n} \Vert^2_2$.
Finally, the contrastive attention mechanism updates $\mathcal{P}$ and further weighs $\mathcal{H}^{v}$ based on $\mathcal{H}^{v,a}$, where $\mathcal{H}^v$ is utilized for the generation process of the final report $\widehat{\mathcal{R}}$.
Similarly, Yan \textit{et al.} \cite{bhi-2022-yan-etal-prior-guided} propose a variational auto-encoder (VAE) based prior encoder to extract prior features from radiographs, and further utilize a prior guided attention layer to weight the prior features and visual features for improving RRG.

\definecolor{lightgray}{gray}{0.97}
\begin{table*}[t]
\centering
\caption{
    The statistics of all radiology datasets with respect to category, the numbers of images, reports, and patients of raw data.
    The number of instances (i.e., radiograph-report pairs) in training, validation, and test sets and the average word-based length of the report in each dataset are also reported.
    ``$^\dagger$'' marks the datasets that randomly split the training, validation, and test sets following the ratio of 7:1:2.
    %
    }
\label{tab: datasets}
\vspace{-0.2cm}
\rowcolors{4}{lightgray}{}
\resizebox{\linewidth}{!}{
\begin{tabular}{llrrrrrrrrr}
\toprule
&& \multicolumn{3}{c}{\textbf{Raw Data}} && \multicolumn{4}{c}{\textbf{Experiment Data}} \\
\cmidrule{3-5} \cmidrule{7-10}
\textbf{Dataset} & \textbf{Category} & Image & Report & Patient && Train \ \ & Val & Test  & Avg. Len. \\
\midrule
CX-CHR$^\dagger$ \cite{nips-2018-hrgr-agent} & Benchmark & 45,598 & 45,598 & 33,236 && 31,919 & 4,559 & 9,120 & - \\
IU X-Ray \cite{iu-xray} & Benchmark & 7,470 & 3,955 & 3,867 && 5,229 & 747 & 1494 & 36.0 \\
MIMIC-CXR \cite{mimic-cxr} & Benchmark & 473,075 & 377,110 & 270,790 && 65,379 & 2,130 & 3,858  & 57.5 \\
MIMIC-CXR-JPG \cite{mimic-cxr-jpg} & Benchmark & 377,110 & 227.835 & 65,379 && 222,758 & 1,808 & 3,269 & 57.5 \\
MIMIC-ABN \cite{mimic-abn} & Benchmark & 38,551 & 38,551 & - && 26,946 & 3,801 & 7,804  & - \\
COV-CTR$^\dagger$ \cite{cov-ctr} & Other & 728 & 728 & 728 && 510 & 72 & 146  & - \\
JLiverCT \cite{nishino-etal-2022-factual} & Other & 882 & 1,083 & 60 && 882 & 127 & 74 & 54.6 \\
PEIR Gross \cite{jing-etal-2018-automatic} & Other & 7,442 & 7,442 & - && - & - & 
- & 12.0 \\
\bottomrule
\end{tabular}
}
\vspace{-0.2cm}
\end{table*}

\smallbreak\noindent\textbf{Memory Networks} \cite{end-to-end-memory-networks}
play as an auxiliary component firstly proposed to leverage external information for question answering,
where a list of memory vectors is used to store different external information, and the vectors are retrieved and weighted based on their importance in answering the questions.
Motivated by this idea, there are studies that apply memory networks to RRG, where
%
the mechanism of memory networks performs the radiograph-report matching process, which is able to significantly facilitate the cross-modal alignment for the RRG task.
Specifically, memory networks are conducted via memory matrix $\mathcal{M}=\{ \mathbf{m}_1 \dots \mathbf{m}_{N_m}\}$, which stores $N_m$ memory vectors to interact with features in different modalities.
Existing approaches 
to adopt cross-modal memory networks for RRG generally contain the following steps.
Given the input radiograph $\mathcal{I}$ and the corresponding report $\mathcal{R}^*$, the visual representation $\mathcal{H}^v$ and textual representation $\mathcal{H}^t$ are firstly computed from $\mathcal{I}$ and $\mathcal{R}^*$.
Then, $\mathcal{H}^v$ and $\mathcal{H}^t$ interact with $\mathcal{M}$ based on two processes, namely memory querying and memory responding, respectively.
Memory querying uses $\mathcal{H}^v$ and $\mathcal{H}^t$ to retrieve the top-$\mathcal{K}$ most related vectors from $\{ \mathbf{m}_1 \dots \mathbf{m}_{N_m} \}$ and obtain their corresponding weights $\{ \omega_1 \dots \omega_{N_m} \}$.
Memory responding applies the weights $\{ \omega_1 \dots \omega_{N_m} \}$ to the memory vectors $\{ \mathbf{m}_1 \dots \mathbf{m}_{N_m} \}$, which results in the responded vector denoted by $\mathbf{r}$.

Owing to the lack of annotated alignment to map cross-modal information for RRG, Qin \textit{et al.} \cite{qin-song-2022-reinforced} further enhance the cross-modal alignment based on Chen \textit{et al.} \cite{chen-etal-2021-cross-modal} with carefully designed reinforcement learning rewards.
It leverages the natural language generation (NLG) metric to guide the cross-modal aligning process.
Given the input radiograph $\mathcal{I}$, the approach follows the standard RRG process in Eq. (\ref{eq: standard_rrg_paradigm}) to generate $\widehat{\mathcal{R}}$, and employs the memory networks to interact with features in different modalities. 
To optimize the overall pipeline, the approach adopts the standard reinforcement learning process in Eq. (\ref{eq: reinforce-reward}) and (\ref{eq: reinforce-gradient}), where the reward $r$ is the improvement on different evaluation metrics (i.e., BLEU \cite{papineni-etal-2002-bleu}) when generating $\widehat{\mathcal{R}}$.
Then, the model is trained to maximize the expected reward so as to enhance the alignment between radiographs and reports.

In leveraging representation weighting, RRG models are able to perform the report generation process better with enhanced cross-modal alignment between radiographs and reports.
The attention mechanism facilitates RRG by modifying the representation with different weights, while the memory networks serve as another form of attentions, which utilizes memory vectors to compute the weights to balance the contributions of different representations.

\subsubsection{Architecture Enhancement}
Architecture enhancement boosts the radiograph-report alignment for RRG by modifying the architectural design of neural networks, where the architecture of the approaches is shown in Figure \ref{sfig: architecture-enhancement}.
In doing so, some studies \cite{you2021aligntransformer, cvpr-2023-kiut, cvpr-2023-metransformer} introduce the task-specific features of radiographs and reports into foundation model architecture such as Transformer \cite{vaswani2017attention}, while the recent study \cite{warm-starting} focuses on offering better initialization for particular model components (e.g., visual encoder or text decoder) with pre-trained model weights.
Specifically, You \textit{et al.} \cite{you2021aligntransformer} align the visual regions of radiographs with specific medical terms using a hierarchical Transformer architecture.
The approach firstly extracts visual features from the input radiograph and term features from medical terms and next concatenates them, where a multi-head self-attention layer is applied to process them, so as to learn the correlation between radiographs and medical terms.
Then, the outputs of different attention layers are sent to a Transformer decoder, and are incorporated through a cross-attention layer with additional learnable parameters that modify the scales, so as to facilitate the report generation process with multiple granular visual-textual alignment.
Existing approaches mainly adopt the standard Transformer architecture for RRG.
For example, Hou \textit{et al.} \cite{miccai-2021-hou-etal-ratchet} combine a pre-trained DenseNet-121 \cite{huang2017densely} with a standard Transformer decoder for RRG.
Huang \textit{et al.} \cite{cvpr-2023-kiut} propose a knowledge-injected U-Transformer architecture for RRG.
The overall architecture consists of two standard Transformer, which contains skip connection between layers of the encoder and decoder to share their internal features.
Besides, extra knowledge is incorporated into each Transformer to enhance the radiograph encoding and report generation processes.
While improving the architecture Transformer is demonstrated to be superior effectiveness, other studies emphasize improving the input of Transformer-based models.
For example, Wang \textit{et al.} \cite{cvpr-2023-metransformer} add additional expert tokens into the input of Transformer-based visual encoder for RRG, where these expert tokens learn the correlations with other visual features and serve as conditions for report generation.
Recently, there are studies that propose multi-task learning model to further improve report generation.
In doing so, Tanwani \textit{et al.} \cite{miccai=2022-tanwani-etal-repsnet} propose a dedicated framework architecture to perform both medical visual question answering (VQA) and report generation, where the encoder aligns radiographs with text descriptions in reports through contrastive learning, and the decoder predicts the report (or answer for VQA) enhanced by related report retrieved from FAISS library\footnote{\url{https://github.com/facebookresearch/faiss}}.
In enhancing the model architecture, RRG models are able to generate elaborated reports with dedicated network architecture.

\begin{figure*}[t]
\centering
\includegraphics[width=1.0\linewidth, trim=0 15 0 0]{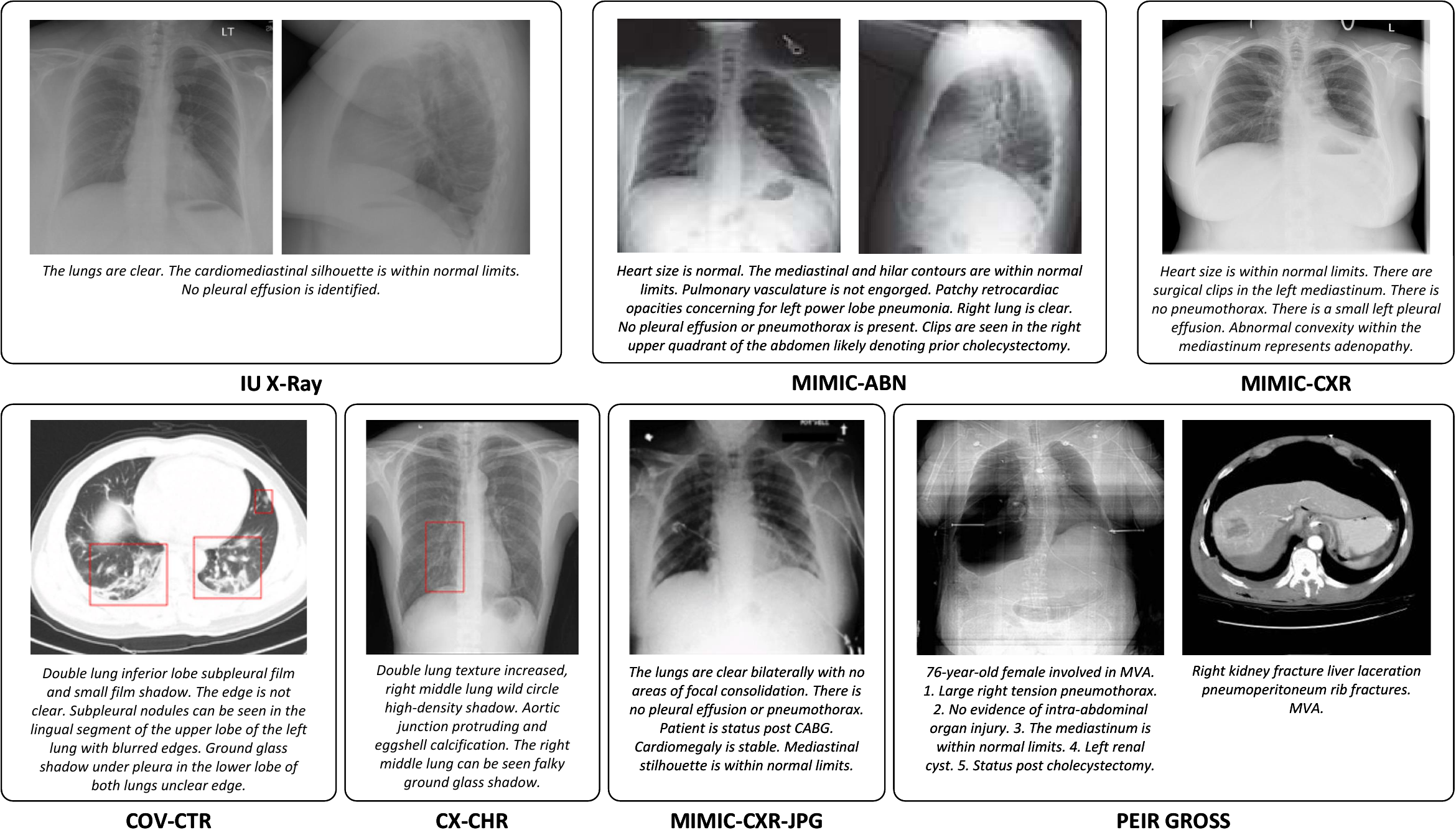}
\caption{
Examples of radiographs and their corresponding reports in IU X-Ray \cite{iu-xray}, MIMIC-CXR \cite{mimic-cxr}, MIMIC-ABN \cite{mimic-abn}, MIMIC-CXR-JPG \cite{mimic-cxr-jpg}, CX-CHR \cite{nips-2018-hrgr-agent}, COV-CTR \cite{cov-ctr}, and PEIR Gross \cite{jing-etal-2018-automatic}, where the radiology reports of CX-CHR and COV-CTR are translated from Chinese into English.
%
The red boxes in radiographs from CX-CHR and COV-CTR are annotated by physicians to highlight the attentive regions that the reports describe.
}
\label{fig: datasets}
\vskip -0.5em
\end{figure*}

\section{Datasets} \label{sec: datasets}

To conduct clinical research with deep learning-based approaches, some studies \cite{nips-2018-hrgr-agent, aaai-2019-chexpert, chexray8-dataset, Liu2020ChestXDet10CX, mimic-cxr, mimic-cxr-jpg, mimic-abn, iu-xray, cov-ctr} manage to construct datasets so as to push forward the development of AI-assisted applications, e.g., mutli-label chest X-ray classification \cite{kipf2017semisupervised, cvpr-2018-tienet, cvpr-2019-multi-label-image-recognition, multi-label-classification-2019, smit2020chexbert}, medical image segmentation \cite{SAM4MIS, samed, anand2023oneshot, ranem2023exploring, Pandey_2023_ICCV, wang2023samocta, zhang2023segment, adapting-sam-for-oct}, doctor-patient communications \cite{wang-etal-2018-coding, tian-etal-2019-chimed, wang-etal-2020-studying, song-etal-2020-summarizing, krishna-etal-2021-generating, michalopoulos-etal-2022-medicalsum}, as well as RRG \cite{jing-etal-2018-automatic, aaai-2019-knowledge, aaai-2020-when, chen-etal-2020-generating, chen-etal-2021-cross-modal, qin-song-2022-reinforced, cvpr-2023-kiut, cvpr-2023-metransformer, kale-etal-2023-replace, kale-etal-2023-kgvl}, and many others.
As for radiograph-driven research, recent studies
construct datasets by collecting chest X-ray images and the corresponding medical records (e.g., radiology reports and medical terminologies) from the hospital, and pre-process the collected data with necessary anonymization such as patient de-identification, so that ensure the quality of these datasets and keep their quantity to meet the requirements of research activities.
We classify prevailing datasets for RRG studies into two main categories, namely benchmark datasets and other datasets, respectively.
Figure \ref{fig: datasets} demonstrates some examples of radiographs and their corresponding reports\footnote{We collect the examples of IU X-Ray, MIMIC-CXR, MIMIC-CXR-JPG from their public datasets, and select the examples of PEIR Gross from \url{http://peir.path.uab.edu/library/}. For CX-CHR and COV-CTR, we present the cases in \url{https://www.researchgate.net/figure/Two-samples-from-CX-CHR-and-COV-CTR-datasets-Red-bounding-boxes-annotated-by-a_fig1_342027618}. Since we have no access to the JLiverCT dataset, the example of JLiverCT is not presented.}, and Table \ref{tab: datasets} reports the statistics of all datasets with respect to the main characteristics, including domain, image number, report number, the number of radiograph-report pairs, patient number, and average length of radiology reports.
Details of the aforementioned datasets in different categories are illustrated in the following texts.

\subsection{Benchmark Datasets}

Benchmark datasets \cite{iu-xray, nips-2018-hrgr-agent, mimic-cxr} are conventionally used by most existing RRG studies to evaluate their model performance and compare it with others.
%
Normally, these datasets contain massive amounts of image-text pairs of radiographs and radiology reports, which is capable of representing the RRG task in terms of its challenging scenarios and ideal expected results.
Moreover, they are able to supply sufficient data resources for existing RRG studies to perform experiments on their approaches, as well as different analyses.
Existing benchmark datasets mainly consist of \textbf{CX-CHR} \cite{nips-2018-hrgr-agent}, \textbf{IU X-Ray} \cite{iu-xray}, and \textbf{MIMIC-CXR} \cite{mimic-cxr}, with details of each specific dataset illustrated as follows.

\smallbreak\noindent\textbf{CX-CHR} \cite{nips-2018-hrgr-agent}
is a benchmark dataset that consists of chest X-Ray images with Chinese reports.
The dataset is collected from a professional medical institution for health checking in China, along with 35,609 patients, 45,598 radiographs, and the corresponding Chinese reports.
Each report contains the conventional findings, and impression sections, along with an additional complains section to reflect syndromes of the corresponding patient from different perspectives.
Particularly, each radiograph is annotated with bounding boxes (see the example in Figure \ref{fig: datasets})
which illustrate the attentive regions that radiologists describe.

\smallbreak\noindent\textbf{IU X-Ray} \cite{iu-xray}
serves as one of the most widely adopted English datasets by existing RRG studies, which is introduced by Indiana University.
To construct the dataset, the authors first collect a series of chest X-Ray images and radiology reports from two large hospital systems within the Indiana Network for Patient Care database.
Then, they remove texts in all radiology reports that reflect the patients' privacy, e.g., patient names, hospital numbers, and complete dates, and filter out the radiographs with potential identifiers such as teeth, partial jaw, jewelry, partial skull, and specific medical devices.
Finally, they manually match these processed radiographs with the corresponding reports, and construct them into radiograph-report pairs for public release, obtaining the final public version of IU X-Ray with 7,470 radiographs and 3,955 reports.
Additionally, the authors use MeSH\footnote{\url{https://www.nlm.nih.gov/mesh/meshhome.html}} to annotate medical terms in all radiology reports, and use RadLex\footnote{\url{http://radlex.org/}} to cover all radiology terminology thatoutsidee of MeSH's purview, resulting in 101 MeSH labels and 76 RadLex labels for all impression and findings sections of radiology reports, respectively.
They also classify all reports into two categories, i.e., normal and abnormal, which contain 2,470 abnormal reports and 1,485 reports, respectively, so that related studies are able to locate these reports according to their demands.
Normally, existing RRG studies follow the data splits proposed by Jing \textit{et al.} \cite{jing-etal-2018-automatic}, which partition IU X-Ray into the training, validation, and testing sets with the ratio of 7:1:2, which results in 5,229, 747, and 1,494 radiograph-report pairs, respectively.

\smallbreak\noindent\textbf{MIMIC-CXR} \cite{mimic-cxr}
is the largest English dataset with chest X-ray radiographs and radiology reports for existing RRG studies, where the dataset is also a commonly used testing benchmark in existing RRG studies.
The dataset is collected from the Beth Israel Deaconess Medical Center (BIDMC) and is designed to support a wide body of research in medicine, including medical image understanding, natural language processing, and clinical decision support.
To construct the dataset, the authors collect 473,075 radiographs and 377,110 reports of 65,379 patients from 2011 to 2016 in the emergency department of BIDMC, and match each radiology report with corresponding radiographs, resulting in 270,790 studies in the public version.
Specifically, radiographs are stored in the format of digital imaging and communications in medicine (DICOM), which contains meta-data associated with one or more radiographs so as to facilitate related studies.
These radiographs are de-identified by removing patient identifiers, with radiologically relevant information such as orientation retained.
For radiology reports, the free-text reports are de-identified by replacing key information (e.g., personal health information (PHI) length) with underscores (i.e., ``\textit{\_ \_ \_}'').
Besides, they employ medical term tagger (e.g., CheXpert \cite{aaai-2019-chexpert}) to annotate medical labels for all radiology reports in MIMIC-CXR, where these labels are classified into 14 categories that are associated with thoracic diseases and support devices.
%
There are studies \cite{mimic-cxr-jpg, mimic-abn} that construct datasets that are extended from MIMIC-CXR, which emphasize improving the data formats or focus on challenging cases of generating reports for chest X-ray radiographs.
Particularly, \textbf{MIMIC-CXR-JPG} \cite{mimic-cxr-jpg} is introduced from BIDMC based on MIMIC-CXR.
The dataset collects 377,110 radiographs and 227,835 reports, and transforms all radiographs from DICOM to JPEG format, so as to facilitate related studies in processing these data.
Moreover, they adopt a more powerful medical term tagger NegBio \cite{negbio-labeler} to annotate medical labels for all radiology reports.
In addition, \textbf{MIMIC-ABN} \cite{mimic-abn} dataset is a subset of MIMIC-CXR with emphasis on reports for abnormal radiographs.
The proposed dataset selects all the abnormal radiographs and corresponding radiology reports from MIMIC-CXR, so as to train neural networks in describing complicated abnormalities of radiographs, and it contains 38,551 radiograph-report pairs in total.

\definecolor{lightgray}{gray}{0.97}
\begin{table*}[t]
\centering
\caption{
An overview of different evaluation metrics with respect to their metric categories, original task, used additional toolkit, and the input type.
}
\label{tab: metrics}
\vspace{-0.2cm}
\rowcolors{4}{lightgray}{}
\setlength{\tabcolsep}{1.2em}
\resizebox{\linewidth}{!}{
\begin{tabular}{lll c cc}
\toprule
\textbf{Metric Category} & & \textbf{Metric} & \textbf{Original Task} & \textbf{Additional Toolkit} & \textbf{Input Type}  \\
\midrule
\cellcolor{white}  & \cellcolor{white}  & BLEU \cite{papineni-etal-2002-bleu} & Translation & - & Report  \\
\cellcolor{white}  \textbf{NLG} & \cellcolor{white}  & METEOR \cite{banerjee-lavie-2005-meteor} & Translation & - & Report \\
\cellcolor{white}  & \cellcolor{white}  & ROUGE \cite{lin-2004-rouge} & Summarization & - & Report  \\
\midrule
\cellcolor{white} & \cellcolor{white}  & Precision & - & CheXpert & Labels \\
\cellcolor{white}  \textbf{CE}  & \cellcolor{white}  & Recall & - & CheXpert & Labels \\
\cellcolor{white}  & \cellcolor{white}  & F1 & - & CheXpert & Labels \\
\midrule
\cellcolor{white} \textbf{SIC}  & \cellcolor{white} & CIDEr \cite{cvpr-2015-cider-metric} & Captioning & - & Report \\
\midrule
\cellcolor{white} \textbf{Embedding} & \cellcolor{white}  & BERTScore \cite{Zhang*2020BERTScore:} & Text Similarity & BERT & Embeddings \\
\midrule
\cellcolor{white}    & \cellcolor{white}  & Factually-oriented \cite{miura-etal-2021-improving} & RRG & - & Entities \\
\cellcolor{white}    & \cellcolor{white}  & MRIQI \cite{aaai-2020-when} & RRG & NegBio, CheXpert & Labels \\
\cellcolor{white}    & \cellcolor{white}  & CO \cite{nishino-etal-2022-factual} & RRG & - & Labels \\
\cellcolor{white}  \multirow{-4}{*}{\cellcolor{white}\textbf{TSF}}    & \cellcolor{white}  & nTKD \cite{iccv-2021-visual-textual} & RRG & - & Labels \\
\bottomrule
\end{tabular}
}
\vspace{-0.2cm}
\end{table*}

\subsection{Other Datasets}

In addition to benchmark datasets, there are some other specific datasets used in existing RRG studies.
These datasets 
are relatively small in scale and diversified compared to benchmark ones,
and usually target on particular applications of RRG, e.g., generating reports for COVID patients' chest radiographs and liver CT images.
Specifically, we mainly introduce three of them, \textbf{COV-CTR} \cite{cov-ctr} \textbf{JLiverCT} \cite{nishino-etal-2022-factual}, and \textbf{PEIR Gross} \cite{jing-etal-2018-automatic} in this paper and illustrate the details of each one in the following texts.

\smallbreak\noindent\textbf{COV-CTR} \cite{cov-ctr}
is a Chinese dataset that focuses on automatically generating radiology reports for CT images of COVID-19 patients.
The dataset is built based on the COVID-CT dataset \cite{zhao2020COVID-CT-Dataset}.
To construct COV-CTR, the authors invite three radiologists from the First Affiliated Hospital of Harbin Medical University to write professional reports for the radiographs of COVID-CT patients.
Finally, the dataset consists of 728 radiographs (349 for COVID-19 and 379 for non-COVID) and their associated Chinese reports.
Although the scale COV-CTR is relatively small compared to other datasets for RRG, the study provides enlightening heuristics for adapting RRG to particular diseases, especially the ones that meet real-time clinical demands.

\smallbreak\noindent\textbf{JLiverCT} \cite{nishino-etal-2022-factual}
is a dataset that focuses on generating radiology reports for liver CT images.
The authors collect radiographs and the corresponding Japanese radiology reports of liver lesions from a hospital, and employ professional radiologists to manually annotate the finding labels in the reports, resulting in 65 categories of finding labels.
For each radiology report, sentences that do not describe the CT images' findings are omitted so as to avoid violating patients' privacy.
Finally, JLiverCT dataset contains 1,083 liver CT images and the corresponding radiology reports.
While most datasets analyze chest X-ray images, the JLiverCT dataset serves as a novel database that emphasizes CT images, which provides insights for the RRG community to address other possible data formats such as MRI imaging.

\smallbreak\noindent\textbf{PEIR Gross} \cite{jing-etal-2018-automatic}
emphasizes producing medical reports for images of gross lesions (i.e., lesion regions visible to human naked eyes).
\textbf{PEIR Gross} consists of medical images and their corresponding medical records from the Pathology Education Informational Resource (PEIR) digital library\footnote{\url{http://peir.path.uab.edu/library/}}.
The dataset contains 7,442 radiographs of different gross lesions, e.g., abdomen, adrenal, chest, gastrointestinal, etc., which are collected from 21 PEIR pathology sub-categories, along with their associated reports.
Since each report in PEIR Gross only contains one descriptive sentence for the corresponding radiograph, which significantly differs from normal reports that are in long contexts, the dataset does not attract much attention from recent studies for RRG.

\section{Evaluation Metrics} \label{sec: evaluation_metrics}

Evaluation metrics are crucial for RRG for providing measurements on the quality of the produced radiology reports from different approaches and ensure a fair comparison among counterparts.
Similar to other AI research, prevailing approaches adopt automatic metrics RRG evaluation through comparing the generated reports with the gold standard references (doctor-written reports).
In general,
metrics for this task are categorized into five types, including natural language generation (NLG), clinical efficacy (CE), standard image captioning (SIC), embedding-based, and task-specific features-based metrics, where NLG metrics and CE metrics are the most widely adopted ones in existing approaches.
Table \ref{tab: metrics} presents an overview of all evaluation metrics with respect to their metric categories, original task, used additional toolkit and the input type.
Details of the aforementioned metrics are illustrated as follows.

\subsection{Natural Language Generation Metrics}
NLG metrics are initially designed for different natural language processing (NLP) tasks, which are adopted for RRG to measure the quality of the generated reports.
Conventional NLG metrics consist of bilingual evaluation understudy (\textbf{BLEU}) \cite{papineni-etal-2002-bleu}, metric for evaluation of translation with explicit ordering (\textbf{METEOR}) \cite{banerjee-lavie-2005-meteor}, and recall-oriented understudy for gisting evaluation (\textbf{ROUGE}) \cite{lin-2004-rouge}.

\smallbreak\noindent\textbf{BLEU} \cite{papineni-etal-2002-bleu}
is introduced for machine translation, which measures the $n$-gram precision of generated tokens.
Given the generated report with $N_r$ tokens and the corresponding gold standard report with $N_{r^*}$ tokens, the BLEU score of $n$-gram $\mathrm{BLEU}$-$n$ is computed by
\begin{equation}
    \mathrm{BLEU}\mbox{-}n = l_{BP} \cdot \exp \left( \sum^{\mathcal{N}}_{i=1} W_i \cdot \log P_i \right)
\end{equation}
where $\mathcal{N}$ refers to the number of $n$-grams, $W_i$ represents the weight of the $i$-th $n$-gram, and $P_i$ denotes the precision of the $i$-th $n$-gram.
Herein, $l_{BP}$ represents the brevity penalty (BP) coefficient to punish the score when the generated report is too short, where $l_{BP} = \exp \left( 
1 - \frac{N_r}{N_{r^*}} \right)$ if $N_r \leq N_{r^*}$ and $l_{BP}=1$ if $N_r > N_{r^*}$.
BLEU-$n$ with $n$ increases from $1$ up to $4$ is usually used for the evaluation of RRG approaches, which measures the accuracy and coherence of the generated report to a certain extent.

\smallbreak\noindent\textbf{METEOR} \cite{banerjee-lavie-2005-meteor}
is designed for machine translation, which computes the recall of matching uni-grams from tokens in produced and gold standard reports according to their exact stemmed form and meaning.
%
Given $N^u$ as the number of uni-grams shared by the generated reports $\widehat{\mathcal{R}}$ and gold standard report $\mathcal{R}^*$, 
the uni-gram precision score $P_u$ and uni-gram recall score $R_u$ are computed by $P_u=\frac{N^u}{N^u_r}$ and $R_r = \frac{N^u}{N^u_{r^*}}$, where $N^u_{r}$ and $N^u_{r^*}$ represent the numbers of uni-grams in $\widehat{\mathcal{R}}$ and $\mathcal{R}^*$, respectively.
Then, the F-score is calculated using harmonic mean through $F = \frac{10\cdot P \cdot R}{R + 9\cdot P}$.
Given the penalty coefficient $p$, the METEOR score is computed by
\begin{equation}
    \mathrm{METEOR} = F \cdot (1 - p)
\end{equation}
where $p=0.5 \cdot \frac{N_c}{N^u}$ and $N_c$ denotes the number of chunks\footnote{A chunk is defined as a set of uni-grams that are adjacent in the generated report and the gold standard report.}.

\smallbreak\noindent\textbf{ROUGE-L} \cite{lin-2004-rouge}
is proposed for summarization and applied for RRG with its variant, e.g., \textbf{ROUGE-L}, which measures the similarity between the generated and gold standard report based on their longest common subsequence (LCS) tokens.
Given the generated report $\widehat{\mathcal{R}}$ with $N_r$ tokens and gold standard report $\mathcal{R}^*$ with $N_{r^*}$ tokens, their LCS tokens are first extracted as $R_{LCS}(\widehat{\mathcal{R}}, \mathcal{R}^*)$.
Then, the precision score $P_{\mathrm{ROUGE-L}}$ and the recall score $R_{\mathrm{ROUGE-L}}$ are computed through $P = \frac{R_{LCS}(\widehat{\mathcal{R}}, \mathcal{R}^*)}{N_r}$ and $R = \frac{R_{LCS}(\widehat{\mathcal{R}}, \mathcal{R}^*)}{N_{r^*}}$, respectively.
Finally, the ROUGE-L score is calculated through
\begin{equation}
    \textrm{ROUGE}\mbox{-}\textrm{L} = \frac{(1 + \beta^2)\cdot R \cdot P}{R+\beta^2 \cdot P}
\end{equation}
where $\beta$ is a hyper-parameter that is usually set to a large number, 
and thus emphasizes the recall score \cite{lin-2004-rouge}.

\subsection{Clinical Efficacy Metrics}
CE metrics are widely utilized in existing approaches to evaluate their model performance in generating factually correct radiology reports.
%
To compute CE metrics, medical labelers (i.e., NegBio \cite{negbio-labeler}, CheXpert \cite{aaai-2019-chexpert}, and MeSH) are firstly employed to label tokens in both generated report $\widehat{\mathcal{R}}$ and the gold standard one $\mathcal{R}^*$, where the obtained labels are related to thoracic diseases and support devices of $14$ categories.
Then, precision, recall, and F1 scores are computed by comparing the medical labels extracted from $\widehat{\mathcal{R}}$ and $\mathcal{R}^*$.
Finally, the aforementioned scores are used as the results of the CE metrics, so as to measure the prediction accuracy of medical labels in the generated report.

\subsection{Standard Image Captioning Metrics}
Standard image captioning (SIC) metrics initially evaluate the quality of generated descriptions for natural images.
Existing RRG approaches \cite{jing-etal-2018-automatic, nips-2018-hrgr-agent, jing-etal-2019-show, aaai-2019-knowledge, aaai-2020-when, cvpr-2021-exploring} mainly adopt consensus-based image description evaluation (\textbf{CIDEr}) \cite{cvpr-2015-cider-metric} following 
the standard image captioning evaluation settings upon the RRG task.
\textbf{CIDEr} measures the quality of generated reports through cosine similarity.
Specifically, CIDEr first calculates the $n$-gram cosine similarities between the produced reports and gold standard ones, obtaining the $n$-gram scores with $n$ ranges from $1$ to $4$.
Then, it computes the mean value over all $n$-gram similarities and uses the resulting value as the final CIDEr score.

\subsection{Embedding-based Metrics}
Embedding-based metrics measure the quality of the generated report through latent distance.
In doing so, feature extractors (e.g., BERT \cite{devlin-etal-2019-bert}) are mainly adopted to project the generated and gold standard reports to the same latent space and compute the distance between their corresponding latent representations as the metric score.
Specifically, current RRG approaches \cite{nishino-etal-2022-factual} utilize BERTScore \cite{Zhang*2020BERTScore:} for the RRG task.
They employ large-scale pre-trained language model (i.e., BERT \cite{devlin-etal-2019-bert}) to help the evaluation of similarity between generated and gold standard reports.
Given the generated report $\widehat{\mathcal{R}}$ and gold standard one $\mathcal{R}^*$, BERTScore uses BERT to encode $\widehat{\mathcal{R}}$ and $\mathcal{R}^*$ into their corresponding latent representations $\mathbf{h}_{\widehat{\mathcal{R}}}$ and $\mathbf{h}_{\mathcal{R}^*}$.
Then, it computes the cosine similarity matrix $\mathcal{M}$ between $\mathbf{h}_{\widehat{\mathcal{R}}}$ and $\mathbf{h}_{\mathcal{R}^*}$, and selects the maximum similarity score $c_{max}$ among $\mathcal{M}$.
Finally, BERTScore uses Inverse Document Frequency (IDF) to provide weights for each different token, and compute the precision, recall, and F1 score based on $c_{max}$, where the F1 score is utilized as the final score of BERTScore.

\subsection{Task-Specific Features-based Metrics}
Task-specific features-based metric aims to utilize the task-specific features (TSF) of RRG to provide comprehensive evaluation from dedicated aspects.
These metrics consist of \textbf{factually-oriented metric} \cite{miura-etal-2021-improving}, medical image report quality index (\textbf{MIRQI}) \cite{aaai-2020-when}, content ordering (\textbf{CO}) \cite{nishino-etal-2022-factual}, and normalized key term distance (\textbf{nTKD}) \cite{iccv-2021-visual-textual}, where the details of them are illustrated in the following texts.

\smallbreak\noindent\textbf{Factually-oriented Metric} \cite{delbrouck-etal-2022-improving}
is introduced to measure the factual correctness and completeness of the generated radiology reports, which contains the exact entity match reward $\mathbf{fact}_{\mathbf{ENT}}$ and the entailing entity match reward $\mathbf{fact}_{\mathbf{ENTNLI}}$.
Specifically, $\mathbf{fact}_{\mathbf{ENT}}$ applies a named entity recognizer (NER) (i.e., Stanza \cite{qi-etal-2020-stanza}) to the generated report $\widehat{\mathcal{R}}$ and the gold standard report $\mathcal{R}^*$, resulting in two sets of entities which are denoted as $E_{\widehat{\mathcal{R}}}$ and $E_{\mathcal{R}^*}$, respectively.
Then, $\mathbf{fact}_{\mathbf{ENT}}$ is computed as the harmonic mean of precision and recall between $E_{\widehat{\mathcal{R}}}$ and $E_{\mathcal{R}^*}$.
Furthermore, $\mathbf{fact}_{\mathbf{ENTNLI}}$ serves as an extension of $\mathbf{fact}_{\mathbf{ENT}}$ with natural language inference (NLI).
Based on the standard process of $\mathbf{fact}_{\mathbf{ENT}}$, $\mathbf{fact}_{\mathbf{ENTNLI}}$ assumes that an entity $e$ of $E_{\widehat{\mathcal{R}}}$ is not considered correct if it presents in $E_{\mathcal{R}^*}$.
To consider $e$ as a valid entity, the sentence $s_{\widehat{\mathcal{R}}}$ containing $e$ must not present a contradiction with its counterpart sentence $s_{\mathcal{R}^*}$ in $\mathcal{R}^*$, where $s_{\mathcal{R}^*}$ refers to the sentence with the highest BERTScore \cite{Zhang*2020BERTScore:} in $\mathcal{R}^*$.
Herein, $\mathbf{fact}_{\mathbf{ENTNLI}}$ utilizes a pre-trained NLI model \cite{miura-etal-2021-improving} to judge whether $s_{\widehat{\mathcal{R}}}$ is a contradiction of $s_{\mathcal{R}^*}$.

\smallbreak\noindent\textbf{MIRQI} \cite{aaai-2020-when}
is proposed to evaluate the quality of reports by measuring the degree of match between the generated reports and the gold standard from the perspectives of the disease keywords and their associated attributes. 
To calculate the metric, medical labeling toolkits are adopted (e.g., NegBio \cite{negbio-labeler} and CheXpert \cite{aaai-2019-chexpert}) to extract two sets of disease words $W_{\widehat{\mathcal{R}}}$ and $W_{\mathcal{R}^*}$ from both generated reports $\widehat{\mathcal{R}}$ and gold standard reports $\mathcal{R}^*$, respectively.
The extracted disease words $W_{\widehat{\mathcal{R}}}$ and $W_{\mathcal{R}^*}$ compose graphs $G_{\widehat{\mathcal{R}}}$ and $G_{\mathcal{R}^*}$ for $\widehat{\mathcal{R}}$ and $\mathcal{R}^*$, where the child nodes of $G_{\widehat{\mathcal{R}}}$ and $G_{\mathcal{R}^*}$ are obtained as the attributes that represent the features of diseases.
Given $G_{\widehat{\mathcal{R}}}$ and $G_{\mathcal{R}^*}$, the precision, recall, and F1 score are computed and used for evaluation by comparing $W_{\widehat{\mathcal{R}}}$ and $W_{\mathcal{R}^*}$ and their associated attributes.

\smallbreak\noindent\textbf{CO} \cite{nishino-etal-2022-factual}
is designed to quantify the consistency of the description order of the generated reports.
It indicates a chronological order of the report contents, and reveals their clinical importance regarding that important findings of a report are likely to be written in the earlier parts.
Given the generated report $\widehat{\mathcal{R}}$ and the gold standard report $\mathcal{R}^*$, the metric first adopts medical labelers (e.g. CheXpert \cite{aaai-2019-chexpert}) to extract finding labels from $\widehat{\mathcal{R}}$ and $\mathcal{R}^*$, which are denoted as $l$ and $l^*$, respectively.
Then, the CO score is computed by the Damerau-Levenshtein distance between $l$ and $l^*$.

\smallbreak\noindent\textbf{nTKD} \cite{iccv-2021-visual-textual}
aims to judge whether sentences in the generated report contain all observed diseases and their detailed descriptive information.
To compute nTKD, the metric firstly employs medical labelers to extract medical labels from the generated report $\widehat{\mathcal{R}}$ and gold standard report $\mathcal{R}^*$, resulting in two sets of labels $l$ and $l^*$, respectively.
Then, the nKTD score $s_{nKTD}$ is computed through Hamming distance $d_{hamming}(\cdot)$ between $l$ and $l^*$, formulated by $s_{nKTD} = \frac{d_{hamming}(l, l^*)}{N_l}$ with $N_l$ as the number of all labels.

\section{Results and Analysis} \label{sec: results_and_analysis}

%
In this section, we summarize the experiment settings of existing RRG studies and conduct a performance comparison on some representative approaches, with analyses of the impact of different approaches according to their model architecture and method categories.
%

\subsection{Experiment Settings}
To assess the effectiveness of RRG, current approaches primarily utilize a range of experiments focusing on various aspects, including quantitative and qualitative evaluations, alongside other specific methods to evaluate its performance.
Quantitative evaluation mainly adopts mainstream metrics (i.e., NLG and CE metrics) for evaluation, which indicates the quality of generated reports with computed scores of different metrics.
%
%
Qualitative evaluation attempts to present visualization or analyses based on the produced results of RRG model.
For example, most existing RRG studies conduct case studies to present the qualitative evaluation, where they select specific radiographs and their corresponding reports from RRG datasets for reference, and directly demonstrate the outputs of different RRG approaches, e.g., radiology reports and medical terms.
Moreover, some attention-based approaches \cite{jing-etal-2018-automatic, iccv-2021-visual-textual, chen-etal-2020-generating, chen-etal-2021-cross-modal, qin-song-2022-reinforced, cvpr-2023-metransformer} visualize the internal attention maps, so as to demonstrate the image-text mappings between specific regions of radiographs and medical terms in the generated reports established by their models.
Particularly, some studies perform method-specific evaluations based on their method categories.
Particularly, knowledge enhancement-based approaches \cite{aaai-2019-knowledge, aaai-2020-when, cvpr-2021-exploring, kale-etal-2023-kgvl, cvpr-2023-dynamic, hou-etal-2023-organ, you-etal-2022-jpg, nishino-etal-2022-factual, kale-etal-2023-replace} evaluate the accuracy of the predicted graph nodes or medical terms for a comprehensive evaluation of their approaches.
RRG studies \cite{cvpr-2021-self-boosting, cvpr-2023-interactive} that employ regional visual features manage to present the visualization of predicted bounding boxes, so as to present the performance of the fine-tuned object detectors of their approaches for RRG.

\subsection{Performance Comparison}
To analyze the impact of different model architectures and method categories, we present a series of performance comparisons upon existing RRG studies.
Table \ref{tab: model-arc-iu-xray} presents the performance comparison on IU X-Ray \cite{iu-xray} among main approaches with different model architectures w.r.t NLG metrics.
Table \ref{tab: mimic-cxr}, \ref{tab: iu-xray}, and \ref{tab: other-datasets} report the performance comparison of main approaches upon MIMIC-CXR \cite{mimic-cxr}, IU X-Ray \cite{iu-xray}, and other datasets (i.e., CX-CHR \cite{jing-etal-2018-automatic}, COV-CTR \cite{cov-ctr}, MIMIC-ABN \cite{mimic-abn} based on the method categories of different studies, respectively.
In the following texts, we first analyze the impacts of different model architectures and summarize the influences of different method categories according to their compared performance on different datasets.

\definecolor{lightgray}{gray}{0.97}
\begin{table}[t]
\centering
\caption{
Comparison of existing approaches on IU X-Ray based on their encoder-decoder architecture, where the background of CNN-LSTM, GCN-LSTM, GCN-Transformer, and Transformer-Transformer architectures are highlighted in grey, red, green, and blue, respectively.
Herein, the \textbf{best} result is highlighted in boldface.
}
\vspace{-0.2cm}
\label{tab: model-arc-iu-xray}
\rowcolors{4}{lightgray}{}
\setlength{\tabcolsep}{0.4em}
\scalebox{0.8}{
\begin{tabular}{lcccccc}
\toprule
 & \multicolumn{6}{c}{\textbf{NLG Metrics}}\\
\cmidrule{2-7}
\textbf{Study} & BLEU-1 & BLEU-2 & BLEU-3 & BLEU-4 & METEOR & ROUGE-L \\
\midrule
\rowcolor{lightgray}
Vinyals \textit{et al.} \cite{vinyals2015tell} & 0.216 & 0.124 & 0.087 & 0.066 & - & 0.306\\
\rowcolor{lightgray}
Rennie \textit{et al.} \cite{rennie2017selfcritical} & 0.224 & 0.129 & 0.089 & 0.068 & - & 0.308 \\
\rowcolor{lightgray}
Lu \textit{et al.} \cite{lu2017knowing} & 0.220 & 0.127 & 0.089 & 0.068 & - & 0.308 \\
\rowcolor{lightgray}
Jing \textit{et al.} \cite{jing-etal-2019-show} & 0.464 & 0.301 & 0.210 & 0.154 & - & 0.362 \\
\rowcolor{lightgray}
Jing \textit{et al.} \cite{jing-etal-2018-automatic} & 0.455 & 0.288 & 0.205 & 0.154 & - & 0.369 \\ 
\rowcolor{lightgray}
Li \textit{et al.} \cite{nips-2018-hrgr-agent} & 0.438 & 0.298 & 0.208 & 0.151 & - & 0.322 \\
\rowcolor{lightgray}
Xue \textit{et al.} \cite{xue2018multimodal} & 0.464 & 0.358 & 0.270 & 0.195 & 0.274 & 0.366 \\
\rowcolor{lightgray}
Harzig \textit{et al.} \cite{harzig2019addressing} & 0.373 & 0.246 & 0.175 & 0.126 & 0.163 & 0.315 \\
\rowcolor{lightgray}
Wang \textit{et al.} \cite{cvpr-2021-self-boosting} & 0.487 & 0.346 & 0.270 & 0.208 & - & 0.359 \\
\rowcolor{lightgray}
Liu \textit{et al.} \cite{liu-etal-2021-contrastive} & 0.492 & 0.314 & 0.222 & 0.169 & 0.193 & 0.381 \\
\rowcolor{lightgray}
Ni \textit{et al.} \cite{iccv-2021-visual-textual} & 0.536 & 0.391 & 0.314 & 0.252 & 0.228 & \textbf{0.448} \\
\rowcolor{lightgray}
Yang \textit{et al.} \cite{yang-etal-2023-joint} & {0.478} & {0.344} & {0.248} & {0.180} & - & {0.398} \\
\rowcolor{lightgray}
\textbf{Avg. score} & 0.397 & 0.265 & 0.194 & 0.146 & 0.215 & 0.349 \\
\midrule
%
%
%
\rowcolor{LightRed}
Zhang \textit{et al.} \cite{aaai-2020-when} & 0.441 & 0.291 & 0.203 & 0.147 & - & 0.367 \\
\midrule
\rowcolor{LightGreen}
Li \textit{et al.} \cite{aaai-2019-knowledge} & 0.455 & 0.288 & 0.205 & 0.154 & - & 0.369 \\ 
\rowcolor{LightGreen}
You \textit{et al.} \cite{you-etal-2022-jpg} & 0.479 & 0.319 & 0.222 & 0.174 & 0.193 & 0.377 \\
\rowcolor{LightGreen}
\textbf{Avg. score} & 0.467 & 0.304 & 0.214 & 0.164 & 0.193 & 0.373 \\
\midrule
\rowcolor{LightBlue}
Chen \textit{et al.} \cite{chen-etal-2020-generating} & 0.470 & 0.304 & 0.219 & 0.165 & - & 0.371 \\ 
\rowcolor{LightBlue}
Nooralahzadeh \textit{et al.} \cite{nooralahzadeh-etal-2021-progressive-transformer} & 0.486 & 0.317 & 0.232 & 0.173 & 0.192 & 0.390 \\
\rowcolor{LightBlue}
Liu \textit{et al.} \cite{cvpr-2021-exploring} & 0.483 & 0.315 & 0.224 & 0.168 & - & 0.376 \\
\rowcolor{LightBlue}
Li \textit{et al.} \cite{cvpr-2023-dynamic} & - & - & - & 0.163 & 0.193 & 0.383 \\
\rowcolor{LightBlue}
Hou \textit{et al.} \cite{hou-etal-2023-organ} & {0.510} & {0.346} & 0.255 & {0.195} & \textbf{0.205} & 0.399 \\
\rowcolor{LightBlue}
Liu \textit{et al.} \cite{liu-etal-2021-contrastive} & 0.492 & 0.314 & 0.222 & 0.169 & 0.193 & 0.381 \\
\rowcolor{LightBlue}
Liu \textit{et al.} \cite{liu-etal-2021-competence} & 0.473 & 0.305 & 0.217 & 0.162 & 0.186 & 0.378 \\
\rowcolor{LightBlue}
Wang \textit{et al.} \cite{cvpr-2021-self-boosting} & 0.487 & {0.346} & {{0.270}} & {{0.208}} & - & 0.359 \\
\rowcolor{LightBlue}
Nooralahzadeh \textit{et al.} \cite{nooralahzadeh-etal-2021-progressive-transformer} & 0.486 & 0.317 & 0.232 & 0.173 & 0.192 & 0.390 \\
\rowcolor{LightBlue}
Chen \textit{et al.} \cite{chen-etal-2021-cross-modal} & 0.475 & 0.309 & 0.222 & 0.170 & 0.191 & 0.375 \\
\rowcolor{LightBlue}
You \textit{et al.} \cite{you2021aligntransformer} & 0.484 & 0.313 & 0.225 & 0.173 & 0.204 & 0.379 \\
\rowcolor{LightBlue}
Hou \textit{et al.} \cite{miccai-2021-hou-etal-ratchet} & 0.232 & - & - & - & 0.101 & 0.240 \\
\rowcolor{LightBlue}
Delbrouck \textit{et al.} \cite{delbrouck-etal-2022-improving} & - & - & - & 0.139 & - & 0.327 \\
\rowcolor{LightBlue}
Wang \textit{et al.} \cite{wang2022cross} & {0.525} & {0.357} & {0.262} & {0.199} & {0.220} & {0.411} \\
\rowcolor{LightBlue}
Qin \textit{et al.} \cite{qin-song-2022-reinforced} & {0.494} & {0.321} & {0.235} & {0.181} & {0.201} & {0.384} \\
\rowcolor{LightBlue}
Yu \textit{et al.} \cite{icbb-2022-yu-etal-clinically-coherent} & {0.457} & {0.305} & {0.216} & {0.171} & - & {0.391} \\
\rowcolor{LightBlue}
Nicolson \textit{et al.} \cite{warm-starting} & {0.473} & {0.304} & {0.224} & {0.175} & 0.200 & {0.376} \\
\rowcolor{LightBlue}
Yan \textit{et al.} \cite{bhi-2022-yan-etal-prior-guided} & {0.482} & {0.313} & {0.232} & {0.181} & 0.203 & {0.381} \\
\rowcolor{LightBlue}
Wang \textit{et al.} \cite{miccai-2022-wang-etal-inclusive} & {0.505} & {0.340} & {0.247} & {0.188} & 0.208 & {0.382} \\
\rowcolor{LightBlue}
Kong \textit{et al.} \cite{miccai-2022-kong-etal-transq} & {0.484} & {0.333} & {0.238} & {0.175} & 0.207 & {0.415} \\
\rowcolor{LightBlue}
Tanwani \textit{et al.} \cite{miccai=2022-tanwani-etal-repsnet} & \textbf{{0.580}} & \textbf{{0.440}} & \textbf{{0.320}} & \textbf{{0.270}} & - & - \\
\rowcolor{LightBlue}
Wang \textit{et al.} \cite{wang-etal-2022-automated} & {0.496} & {0.319} & {0.241} & {0.175} & - & {0.377} \\
\rowcolor{LightBlue}
Wang \textit{et al.} \cite{Wang2022SelfAG} & 0.505 & 0.345 & 0.243 & 0.176 & 205 & 0.396 \\ 
\rowcolor{LightBlue}
Huang \textit{et al.} \cite{cvpr-2023-kiut} & {{0.525}} & {{0.360}} & 0.251 & 0.185 & \textbf{0.242} & {0.409} \\
\rowcolor{LightBlue}
Wang \textit{et al.} \cite{cvpr-2023-metransformer} & 0.483 & 0.322 & 0.228 & 0.172 & 0.192 & 0.380 \\
\rowcolor{LightBlue}
\textbf{Avg. score} & {0.480} & {0.329} & {0.223} & {0.200} & 0.196 & 0.414 \\
\bottomrule
\end{tabular}
}
\vspace{-0.2cm}
\end{table}

\subsubsection{Comparison of Model Architecture}
The model architectures of existing RRG studies are categorized into three main groups, including CNN-LSTM architecture, GCN-LSTM architecture, GCN-Transformer architecture, and Transformer-Transformer architecture, whose details are illustrated as follows.
Specifically, CNN-LSTM architecture is adopted by many approaches \cite{jing-etal-2018-automatic, cvpr-2021-self-boosting, iccv-2021-visual-textual}. 
It uses CNN or its variants to encode radiographs into latent features and utilizes LSTM to decode sentences in the final report.
These approaches provide heuristics for RRG, and contain significant room for performance improvement.
Some studies \cite{aaai-2019-knowledge, aaai-2020-when, cvpr-2021-exploring, kale-etal-2023-kgvl, cvpr-2023-dynamic, hou-etal-2023-organ} manage to introduce knowledge graph to facilitate the report generation process, which motivates the requirements of processing graph-based information for RRG.
Therefore, GCN-based encoders are proposed and lead to improvements on CNN-LSTM-based approaches, demonstrating the effectiveness of knowledge graphs for RRG, where such encoders help the model build relationships between diseases and organs through encoding the corresponding nodes and edges of the knowledge graph for the LSTM decoder, thereby performing a more effective encoding process compared to CNN-based encoders.
With the rising of Transformer-based architecture, recent studies \cite{chen-etal-2021-cross-modal, cvpr-2021-exploring, nooralahzadeh-etal-2021-progressive-transformer, yan-etal-2021-weakly-supervised, qin-song-2022-reinforced, you-etal-2022-jpg, nishino-etal-2022-factual, cvpr-2023-interactive, cov-ctr, cvpr-2023-dynamic, hou-etal-2023-organ, kale-etal-2023-replace, cvpr-2023-kiut, cvpr-2023-metransformer} employ Transformer for RRG, which is first utilized to improve the text decoder, and then employed to enhance both encoders and decoders.
Specifically, GCN-Transformer architecture brings further improvements upon GCN-LSTM architecture, with Transformer-based decoders to handle the report generation process more effectively.
Later, Transformer is also conducted for encoders and obtains the best performance over existing model architectures, where the multi-head attention mechanism in Transformer demonstrates its superiority for both encoding radiographs and generating radiology reports.
This observation demonstrates that the multi-head attention mechanism currently serves as the best solution to model the long report generation process, where the self-attention layers learn the internal correlation of regions in radiographs as well as words in radiology reports.
%
%

\definecolor{lightgray}{gray}{0.97}
\begin{table*}[t]
\centering
\caption{
Results of representative RRG models on MIMIC-CXR with respect to NLG and CE metrics, where image captioning approaches, visual-only approaches, textual-only approaches, and cross-modal approaches are marked by gray, red, green, and blue backgrounds, respectively.
%
%
%
}
\label{tab: mimic-cxr}
\setlength{\tabcolsep}{0.7em}
\scalebox{1.1}{
\begin{tabular}{lccccccccccc}
\toprule
 & \multicolumn{6}{c}{\textbf{NLG Metrics}} & & \multicolumn{3}{c}{\textbf{CE Metrics}} \\
\cmidrule{2-7} \cmidrule{9-11}
\textbf{Study} & BLEU-1 & BLEU-2 & BLEU-3 & BLEU-4 & METEOR & ROUGE-L & & Precision & Recall & F1 \\
\midrule
\rowcolor{lightgray}
Vinyals \textit{et al.} \cite{vinyals2015tell} & 0.299 & 0.184 & 0.121 & 0.084 & 0.124 & 0.263 & & 0.249 & 0.203 & 0.204 \\ 
\rowcolor{lightgray}
Rennie \textit{et al.} \cite{rennie2017selfcritical} & 0.325 & 0.203 & 0.136 & 0.096 & 0.134 & 0.276 && 0.322 & 0.239 & 0.249 \\
\rowcolor{lightgray}
Lu \textit{et al.} \cite{lu2017knowing} & 0.299 & 0.185 & 0.124 & 0.088 & 0.118 & 0.266 && 0.268 & 0.186 & 0.181 \\
\rowcolor{lightgray}
Anderson \textit{et al.} \cite{anderson2018bottomup} & 0.317 & 0.195 & 0.130 & 0.092 & 0.128 & 0.267 && 0.320 & 0.231 & 0.238 \\
\rowcolor{lightgray}
\textbf{Avg. score} & 0.310 & 0.192 & 0.128 & 0.090 & 0.126 & 0.268 && 0.290 & 0.215 & 0.218 \\
\midrule
\rowcolor{LightRed}
Tanida \textit{et al.} \cite{cvpr-2023-interactive}  & 0.373 & 0.249 & {0.175} & {0.126} & {0.168} & 0.264 && {0.461} & {0.475} & {0.447} \\
\rowcolor{LightRed}
Wang \textit{et al.} \cite{Wang2022SelfAG}  & 0.363 & 0.235 & {0.164} & {0.118} & {0.136} & 0.301 && - & - & - \\
\rowcolor{LightRed}
\textbf{Avg. score} & 0.368 & 0.242 & 0.170 & 0.122 & 0.152 & 0.283 && - & - & - \\
\midrule
\rowcolor{LightGreen}
Chen \textit{et al.} \cite{chen-etal-2020-generating} & 0.353 & 0.218 & 0.145 & 0.103 & 0.142 & 0.277 && 0.333 & 0.273 & 0.276 \\
\rowcolor{LightGreen}
Nooralahzadeh \textit{et al.} \cite{nooralahzadeh-etal-2021-progressive-transformer} & 0.378 & 0.232 & 0.154 & 0.107 & 0.145 & 0.272 && 0.240 & 0.428 & 0.308 \\
\rowcolor{LightGreen}
Liu \textit{et al.} \cite{cvpr-2021-exploring} & 0.360 & 0.224 & 0.149 & 0.106 & 0.149 & 0.284 && - & - & - \\
\rowcolor{LightGreen}
Nishino \textit{et al.} \cite{nishino-etal-2022-factual} & - & - & - & 0.122 & - & 0.168 && - & - & - \\
\rowcolor{LightGreen}
Dalla Serra \textit{et al.} \cite{dalla-serra-etal-2022-multimodal} & 0.363 & 0.245 & {0.178} & \textbf{0.136} & 0.161 & \textbf{{0.313}} && 0.428 & 0.459 & 0.443 \\
\rowcolor{LightGreen}
Li \textit{et al.} \cite{cvpr-2023-dynamic} & - & - & - & 0.109 & 0.150 & 0.284 && - & - & - \\
\rowcolor{LightGreen}
Hou \textit{et al.} \cite{hou-etal-2023-organ} & {0.407} & {0.256} & 0.172 & 0.123 & {0.162} & 0.293 && 0.416 & 0.418 & 0.385 \\
\rowcolor{LightGreen}
\textbf{Avg. score} & 0.372 & 0.235 & {0.160} & 0.115 & {0.151} & {0.270} && {0.329} & {0.395} & {0.353} \\
\midrule
\rowcolor{LightBlue}
Nishino \textit{et al.} \cite{nishino-etal-2020-reinforcement} & 0.363 & 0.231 & 0.163 & 0.121 & - & 0.267 && - & - & - \\ 
\rowcolor{LightBlue}
Chen \textit{et al.} \cite{chen-etal-2021-cross-modal} & 0.353 & 0.218 & 0.148 & 0.106 & 0.142 & 0.278 && 0.334 & 0.275 & 0.278 \\ 
\rowcolor{LightBlue}
You \textit{et al.} \cite{you2021aligntransformer} & 0.378 & 0.235 & 0.156 & 0.112 & 0.158 & 0.283 && - & - & - \\
\rowcolor{LightBlue}
Yan \textit{et al.} \cite{yan-etal-2021-weakly-supervised} & 0.373 & - & - & 0.107 & 0.144 & 0.274 && 0.385 & 0.274 & 0.294 \\
\rowcolor{LightBlue}
Ni \textit{et al.} \cite{iccv-2021-visual-textual} & 0.372 & 0.241 & 0.168 & 0.123 & \textbf{0.190} & 0.335 && - & - & - \\
\rowcolor{LightBlue}
Delbrouck \textit{et al.} \cite{delbrouck-etal-2022-improving} & - & - & - & 0.116 & - & 0.259 && - & - & - \\
\rowcolor{LightBlue}
Qin \textit{et al.} \cite{qin-song-2022-reinforced} & 0.381 & 0.232 & 0.155 & 0.109 & 0.151 & {0.287} && 0.342 & 0.294 & 0.292 \\
\rowcolor{LightBlue}
Wang \textit{et al.} \cite{wang2022cross} & 0.344 & 0.215 & 0.146 & 0.105 & 0.138 & {0.279} && - & - & - \\
\rowcolor{LightBlue}
Yu \textit{et al.} \cite{icbb-2022-yu-etal-clinically-coherent} & 0.347 & 0.235 & 0.149 & 0.106 & - & {0.280} && 0.447 & 0.593 & 0.503 \\
\rowcolor{LightBlue}
Nicolson \textit{et al.} \cite{warm-starting} & 0.393 & 0.248 & 0.171 & 0.127 & 0.155 & {0.286} && 0.367 & 0.418 & 0.391 \\
\rowcolor{LightBlue}
Yan \textit{et al.} \cite{bhi-2022-yan-etal-prior-guided} & 0.356 & 0.222 & 0.151 & 0.111 & 0.140 & {0.280} && 0.353 & 0.310 & 0.297 \\
\rowcolor{LightBlue}
Wang \textit{et al.} \cite{miccai-2022-wang-etal-inclusive} & 0.395 & 0.253 & 0.170 & 0.121 & 0.147 & {0.284} && - & - & - \\
\rowcolor{LightBlue}
Wang \textit{et al.} \cite{miccai-2022-wang-etal-semantic-assisted} & 0.413 & \textbf{0.266} & \textbf{0.186} & \textbf{0.136} & 0.168 & {0.286} && \textbf{0.482} & \textbf{0.563} & \textbf{0.519} \\
\rowcolor{LightBlue}
Kong \textit{et al.} \cite{miccai-2022-kong-etal-transq} & \textbf{0.423} & 0.261 & 0.171 & 0.116 & 0.170 & {0.298} && - & - & - \\
\rowcolor{LightBlue}
Wang \textit{et al.} \cite{wang-etal-2022-automated} & {0.351} & 0.223 & 0.157 & 0.118 & - & {0.287} && - & - & - \\
\rowcolor{LightBlue}
Yang \textit{et al.} \cite{yang-etal-2023-joint} & 0.362 & 0.251 & 0.188 & 0.143 & - & {0.326} && - & - & - \\
\rowcolor{LightBlue}
Huang \textit{et al.} \cite{cvpr-2023-kiut} & {0.393} & 0.243 & 0.159 & 0.113 & 0.160 & 0.285 && 0.371 & 0.318 & 0.321 \\
\rowcolor{LightBlue}
Wang \textit{et al.} \cite{cvpr-2023-metransformer} & 0.386 & {0.250} & 0.169 & 0.124 & 0.152 & 0.291 && 0.364 & 0.309 & 0.311 \\
\rowcolor{LightBlue}
\textbf{Avg. score} & {0.378} & {0.239} & 0.161 & {0.116} & 0.155 & {0.285} && 0.383 & 0.373 & 0.356 \\
\bottomrule
\end{tabular}
}
\vspace{-0.2cm}
\end{table*}

\subsubsection{Comparison of Method Categories}
\begin{figure*}[t]
\centering
\includegraphics[width=1.0\linewidth, trim=0 10 0 0]{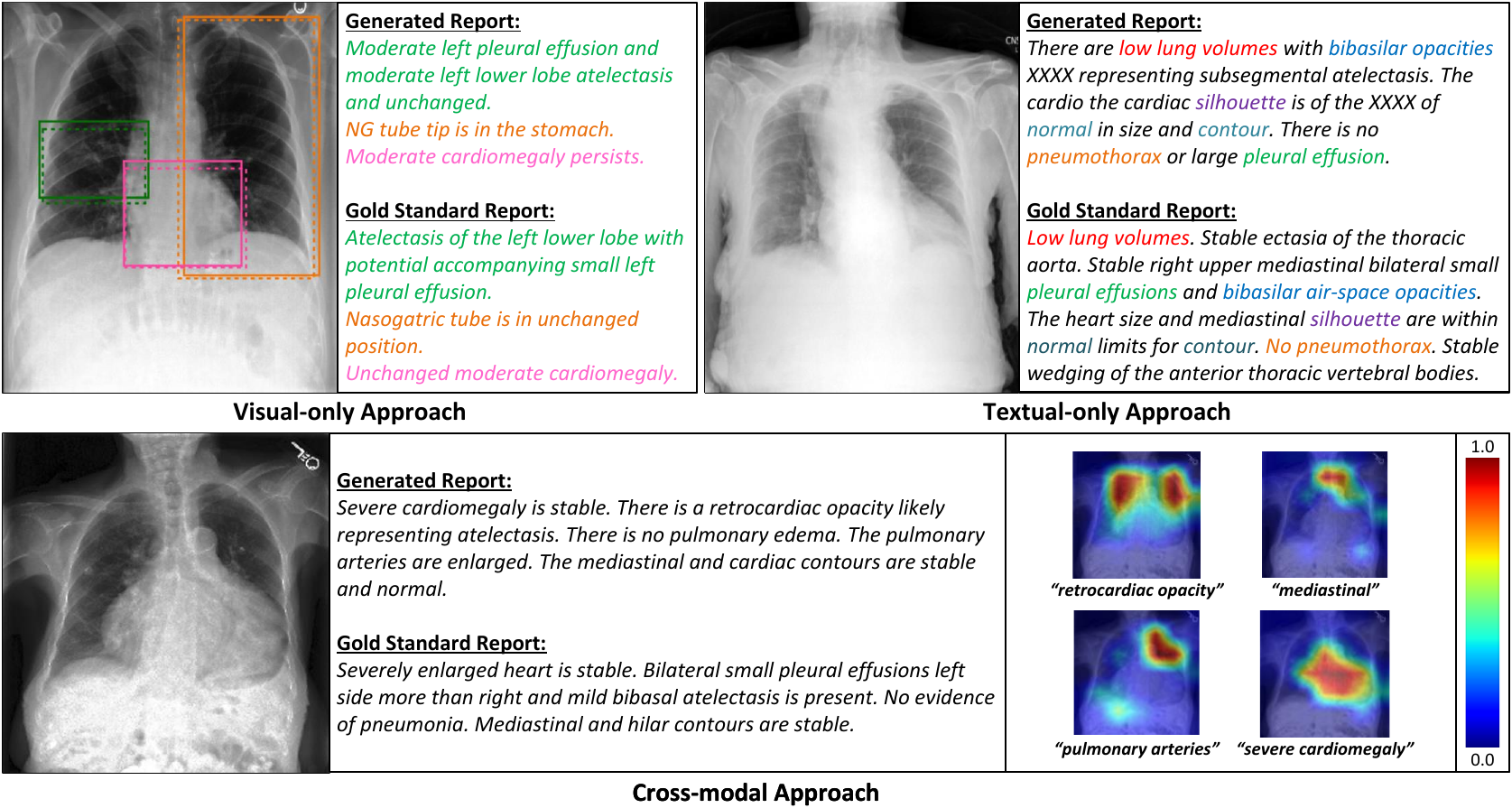}
\caption{
Illustrations of the reports generated by visual-only, textual-only, and cross-modal approaches according to their input radiographs, where the gold standard reports are presented for reference.
Herein, results of visual-only, textual-only, and cross-modal approaches are produced by Tanida \textit{et al.} \cite{cvpr-2023-interactive}, You \textit{et al.} \cite{you-etal-2022-jpg}, and Chen \textit{et al.} \cite{chen-etal-2021-cross-modal}, respectively.
For visual-only approach, detected regions and the corresponding descriptive sentences are highlighted in the same color.
For textual-only approach, medical terms shared by the generated report and the gold standard report are highlighted in the same colors.
For cross-modal approach, we demonstrate the visualization of image-text mappings between particular regions of the radiograph and words/phrases from its generated report, with the color spectrum indicating the value of weight ranging from $[0,1]$.
}
\label{fig: case-study}
\end{figure*}

To analyze the impact of method categories in existing studies, we first illustrate the impacts of each specific category, and then compare them across different categories to discuss the effect brought by different modality enhancement.
For reference, we report the performance of image captioning approaches on RRG datasets.
In Figure \ref{fig: case-study}, we demonstrate some examples of generated reports by visual-only, textual-only, and cross-modal approaches with respect to different characteristics.
Details are introduced in the following texts.

\definecolor{lightgray}{gray}{0.97}
\begin{table}[t]
\centering
\caption{
Comparisons of existing RRG approaches on the test sets of IU X-Ray with respect to NLG metrics, where the image captioning approaches, textual-only approaches, and cross-modal approaches are highlighted in gray, green, and blue, backgrounds, respectively.
%
}
\label{tab: iu-xray}
\rowcolors{4}{lightgray}{}
\setlength{\tabcolsep}{0.4em}
\scalebox{0.8}{
\begin{tabular}{lcccccc}
\toprule
 & \multicolumn{6}{c}{\textbf{NLG Metrics}}\\
\cmidrule{2-7}
\textbf{Study} & BLEU-1 & BLEU-2 & BLEU-3 & BLEU-4 & METEOR & ROUGE-L \\
\midrule
\rowcolor{lightgray}
Vinyals \textit{et al.} \cite{vinyals2015tell} & 0.216 & 0.124 & 0.087 & 0.066 & - & 0.306\\
\rowcolor{lightgray}
Rennie \textit{et al.} \cite{rennie2017selfcritical} & 0.224 & 0.129 & 0.089 & 0.068 & - & 0.308 \\
\rowcolor{lightgray}
Lu \textit{et al.} \cite{lu2017knowing} & 0.220 & 0.127 & 0.089 & 0.068 & - & 0.308 \\
\rowcolor{lightgray}
\textbf{Avg. score} & 0.220 & 0.127 & 0.088 & 0.067 & - & 0.307 \\ 
\midrule
\rowcolor{LightRed}
Wang \textit{et al.} \cite{Wang2022SelfAG} & 0.505 & 0.345 & 0.243 & 0.176 & 205 & 0.396 \\ 
\midrule
\rowcolor{LightGreen}
Li \textit{et al.} \cite{aaai-2019-knowledge} & 0.455 & 0.288 & 0.205 & 0.154 & - & 0.369 \\ 
\rowcolor{LightGreen}
Jing \textit{et al.} \cite{jing-etal-2019-show} & 0.464 & 0.301 & 0.210 & 0.154 & - & 0.362 \\
\rowcolor{LightGreen}
Harzig \textit{et al.} \cite{harzig2019addressing} & 0.373 & 0.246 & 0.175 & 0.126 & 0.163 & 0.315 \\
\rowcolor{LightGreen}
Zhang \textit{et al.} \cite{aaai-2020-when} & 0.441 & 0.291 & 0.203 & 0.147 & - & 0.367 \\
\rowcolor{LightGreen}
Chen \textit{et al.} \cite{chen-etal-2020-generating} & 0.470 & 0.304 & 0.219 & 0.165 & - & 0.371 \\ 
\rowcolor{LightGreen}
Nooralahzadeh \textit{et al.} \cite{nooralahzadeh-etal-2021-progressive-transformer} & 0.486 & 0.317 & 0.232 & 0.173 & 0.192 & 0.390 \\
\rowcolor{LightGreen}
Liu \textit{et al.} \cite{cvpr-2021-exploring} & 0.483 & 0.315 & 0.224 & 0.168 & - & 0.376 \\
\rowcolor{LightGreen}
You \textit{et al.} \cite{you-etal-2022-jpg} & 0.479 & 0.319 & 0.222 & 0.174 & 0.193 & 0.377 \\
\rowcolor{LightGreen}
Li \textit{et al.} \cite{cvpr-2023-dynamic} & - & - & - & 0.163 & 0.193 & 0.383 \\
\rowcolor{LightGreen}
Hou \textit{et al.} \cite{hou-etal-2023-organ} & {0.510} & {0.346} & 0.255 & {0.195} & 0.205 & 0.399 \\
\rowcolor{LightGreen}
\textbf{Avg. score} & 0.474 & 0.310 & 0.221 & 0.166 & {0.196} & {0.377}  \\
\midrule
\rowcolor{LightBlue}
Jing \textit{et al.} \cite{jing-etal-2018-automatic} & 0.455 & 0.288 & 0.205 & 0.154 & - & 0.369 \\ 
\rowcolor{LightBlue}
Li \textit{et al.} \cite{nips-2018-hrgr-agent} & 0.438 & 0.298 & 0.208 & 0.151 & - & 0.322 \\
\rowcolor{LightBlue}
Liu \textit{et al.} \cite{liu-etal-2021-contrastive} & 0.492 & 0.314 & 0.222 & 0.169 & 0.193 & 0.381 \\
\rowcolor{LightBlue}
Liu \textit{et al.} \cite{liu-etal-2021-competence} & 0.473 & 0.305 & 0.217 & 0.162 & 0.186 & 0.378 \\
\rowcolor{LightBlue}
Wang \textit{et al.} \cite{cvpr-2021-self-boosting} & 0.487 & {0.346} & {{0.270}} & {{0.208}} & - & 0.359 \\
\rowcolor{LightBlue}
Chen \textit{et al.} \cite{chen-etal-2021-cross-modal} & 0.475 & 0.309 & 0.222 & 0.170 & 0.191 & 0.375 \\
\rowcolor{LightBlue}
You \textit{et al.} \cite{you2021aligntransformer} & 0.484 & 0.313 & 0.225 & 0.173 & 0.204 & 0.379 \\
\rowcolor{LightBlue}
Hou \textit{et al.} \cite{miccai-2021-hou-etal-ratchet} & 0.232 & - & - & - & 0.101 & 0.240 \\
\rowcolor{LightBlue}
Ni \textit{et al.} \cite{iccv-2021-visual-textual} & 0.536 & 0.391 & 0.314 & {0.252} & 0.228 & \textbf{0.448} \\
\rowcolor{LightBlue}
Delbrouck \textit{et al.} \cite{delbrouck-etal-2022-improving} & - & - & - & 0.139 & - & 0.327 \\
\rowcolor{LightBlue}
Qin \textit{et al.} \cite{qin-song-2022-reinforced} & {0.494} & {0.321} & {0.235} & {0.181} & {0.201} & {0.384} \\
\rowcolor{LightBlue}
Wang \textit{et al.} \cite{wang2022cross} & {0.525} & {0.357} & {0.262} & {0.199} & {0.220} & {0.411} \\
\rowcolor{LightBlue}
Yu \textit{et al.} \cite{icbb-2022-yu-etal-clinically-coherent} & {0.457} & {0.305} & {0.216} & {0.171} & - & {0.391} \\
\rowcolor{LightBlue}
Nicolson \textit{et al.} \cite{warm-starting} & {0.473} & {0.304} & {0.224} & {0.175} & 0.200 & {0.376} \\
\rowcolor{LightBlue}
Yan \textit{et al.} \cite{bhi-2022-yan-etal-prior-guided} & {0.482} & {0.313} & {0.232} & {0.181} & 0.203 & {0.381} \\
\rowcolor{LightBlue}
Kong \textit{et al.} \cite{miccai-2022-kong-etal-transq} & {0.484} & {0.333} & {0.238} & {0.175} & 0.207 & {{0.415}} \\
\rowcolor{LightBlue}
Wang \textit{et al.} \cite{miccai-2022-wang-etal-inclusive} & {0.505} & {0.340} & {0.247} & {0.188} & 0.208 & {0.382} \\
\rowcolor{LightBlue}
Tanwani \textit{et al.} \cite{miccai=2022-tanwani-etal-repsnet} & \textbf{{0.580}} & \textbf{{0.440}} & \textbf{{0.320}} & \textbf{{0.270}} & - & - \\
\rowcolor{LightBlue}
Wang \textit{et al.} \cite{wang-etal-2022-automated} & {0.496} & {0.319} & {0.241} & {0.175} & - & {0.377} \\
\rowcolor{LightBlue}
Yang \textit{et al.} \cite{yang-etal-2023-joint} & {0.478} & {0.344} & {0.248} & {0.180} & - & {0.398} \\
\rowcolor{LightBlue}
Huang \textit{et al.} \cite{cvpr-2023-kiut} & {{0.525}} & {{0.360}} & 0.251 & 0.185 & \textbf{0.242} & {0.409} \\
\rowcolor{LightBlue}
Wang \textit{et al.} \cite{cvpr-2023-metransformer} & 0.483 & 0.322 & 0.228 & 0.172 & 0.192 & 0.380 \\
\rowcolor{LightBlue}
\textbf{Avg. score} & {0.478} & {0.331} & {0.241} & {0.183} & 0.198 & 0.374 \\
\bottomrule
\end{tabular}
}
\vspace{-0.2cm}
\end{table}

\smallbreak\noindent\textbf{Visual-only Approaches}
mostly adopt global visual features for RRG, where several studies employ regional visual features or global-regional aggregated ones for further enhancement.
By comparing whether using regional visual features for RRG, Tanida \textit{et al.} \cite{cvpr-2023-interactive} that employ regional visual features obtain improvements over other approaches using global visual features (e.g., Wang \textit{et al.} \cite{cvpr-2023-metransformer}), suggesting the effectiveness of utilizing regional visual features to provide fine-grained visual information for the report generation process.
Notably, this approach obtains significantly improved performance on CE metrics over approaches that use global visual features, which indicates that regional visual features are able to facilitate the overall pipeline in matching particular regions in radiographs to medical terms in radiology reports, as is demonstrated in Figure \ref{fig: case-study}.
Also, global-regional aggregated features also demonstrate its effectiveness in improving RRG, where Li \textit{et al.} \cite{cov-ctr} reveal better performance over Li \textit{et al.} \cite{aaai-2019-knowledge} on CX-CHR. 
This observation verifies the validity of considering both global and regional views of radiographs for RRG, which is similar to the report-writing process of physicians.
Nevertheless, the number of visual-only approaches is still limited, while existing studies mainly focus on employing textual information and enhanced radiograph-report alignment to assist RRG, so the paradigm of visual-only approaches still has large potentials for further enhancement.

\smallbreak\noindent\textbf{Textual-only Approaches}
mainly enhance RRG through knowledge enhancement, report structuralization, and progressive report generation, with several observations as follows.
Knowledge enhancement-based approaches \cite{aaai-2019-knowledge, aaai-2020-when, cvpr-2021-exploring, you-etal-2022-jpg, nishino-etal-2022-factual, kale-etal-2023-kgvl, cvpr-2023-dynamic, hou-etal-2023-organ, kale-etal-2023-replace} play a dominant role in existing RRG studies.
They employ textual information (e.g., medical terms, entities, and relations, as well as knowledge graph) to facilitate RRG.
Particularly, some studies that adopt medical terms (e.g., You \textit{et al.} \cite{you-etal-2022-jpg}) demonstrate significant improvement compared to vanilla image captioning approaches, indicating that the report generation process is effectively guided through disease classification and medical terms prediction.
Knowledge graph-based approaches (e.g., Zhang \textit{et al.} \cite{aaai-2020-when}, Li \textit{et al.} \cite{cvpr-2023-dynamic}, and Hou \textit{et al.} \cite{hou-etal-2023-organ}) are one of the most mainstream approaches since the employment of deep learning in RRG, where these approaches also obtain competitive performance with medical term-based ones.
Besides, approaches \cite{aaai-2019-knowledge, dalla-serra-etal-2022-multimodal} that adopt structuralized information (e.g., report templates) to facilitate RRG demonstrate competitive performance as well as knowledge-enhanced ones.
Notably, the improvements of these RRG models over general image captioning approaches are particularly on BLEU-3 and BLEU-4 metrics, which indicates that the document-level guidance is able to pilot the text decoder in generating more coherent reports.
Moreover, the progressive report generation-based approaches (e.g., Nooralahzadeh \textit{et al.} \cite{nooralahzadeh-etal-2021-progressive-transformer}) also outperform image captioning approaches, with the two-stage generation paradigm indicating the effectiveness of generating reports in long contexts.
Nevertheless, the performance gain of progressive report generation is less significant than other categories, suggesting that the impact of different text guidance upon the generation process varies according to the granularity level of guidance, where word-level information (e.g., medical terms and knowledge graphs) provides fine-grained guidance in generating descriptive texts, and document-level information (e.g., report template, related report, and intermediate result) puts emphasis on the construction of the entire radiology report.

\smallbreak\noindent\textbf{Cross-modal Approaches}
generally
all obtain competitive performance owing to their privileging model designs, including reinforcement learning, representation weighting, and architecture enhancement ones, respectively.
In detail,
reinforcement learning-based approaches \cite{jing-etal-2019-show, nishino-etal-2020-reinforcement, cvpr-2021-self-boosting, liu-etal-2021-competence, delbrouck-etal-2022-improving, qin-song-2022-reinforced} that optimize the training objectives improve the RRG performance through carefully designed strategies such as reinforcement learning rewards and additional supervision signals, which obtain significant improvements over the image captioning approaches with standard cross-entropy loss.
This observation illustrates that there are still improvements for RRG to design dedicated objectives, so that the cross-modal alignment radiographs and reports are enhanced to facilitate RRG.
Representation weighting-based approaches \cite{jing-etal-2018-automatic, chen-etal-2020-generating, liu-etal-2021-contrastive, cvpr-2021-exploring, iccv-2021-visual-textual, chen-etal-2021-cross-modal, qin-song-2022-reinforced, cvpr-2023-kiut, cvpr-2023-metransformer} also outperform previous image captioning approaches, where the attention mechanisms \cite{jing-etal-2018-automatic, iccv-2021-visual-textual, liu-etal-2021-contrastive} learn to weight the representation in different modalities, and the memory networks are optimized to establish the image-text matching between radiographs and reports.
With particular model architecture designed for RRG, the designed RRG models (e.g., \cite{cvpr-2023-kiut} and \cite{cvpr-2023-metransformer}) show promising performance, where RRG models are more capable of modeling the long text generation process than image captioning ones.
The aforementioned observations illustrate the effectiveness of improving the cross-modal alignment based on whether additional modifications upon model architecture and optimization strategies are needed, which is able to provide references to adapt existing RRG studies to other applications in the future based on their application-specific characteristics.

\definecolor{lightgray}{gray}{0.97}
\begin{table}[t]
\centering
\caption{
Comparisons of existing RRG approaches on the test sets of CX-CHR,  COV-CTR, and JLiverCT datasets, with respect to NLG metrics.
%
}
\label{tab: other-datasets}
\rowcolors{4}{lightgray}{}
\setlength{\tabcolsep}{0.4em}
\scalebox{0.82}{
\begin{tabular}{llcccccc}
\toprule
 & & \multicolumn{5}{c}{\textbf{NLG Metrics}} & \\
\cmidrule{3-7}
\textbf{Study} & \textbf{Dataset} & BLEU-1 & BLEU-2 & BLEU-3 & BLEU-4 & ROUGE-L  \\
\midrule
    Li \textit{et al.} \cite{aaai-2019-knowledge} & CX-CHR & 0.673 & 0.588 & 0.532 & 0.473 & 0.618 \\ 
    Li \textit{et al.} \cite{cov-ctr} & CX-CHR & \textbf{0.686} & \textbf{0.608} & \textbf{0.558} & \textbf{0.523} & \textbf{0.641}  \\
    \midrule
    Wang \textit{et al.} \cite{cvpr-2021-self-boosting} & COV-CTR & \textbf{0.810} & \textbf{0.766} & \textbf{0.721} & \textbf{0.679} & \textbf{0.790}  \\
    Li \textit{et al.} \cite{cov-ctr} & COV-CTR & 0.712 & 0.659 & 0.611 & 0.570 & 0.746  \\
    \midrule
    Yan \textit{et al.} \cite{yan-etal-2021-weakly-supervised} & MIMIC-ABN & 0.256 & - & - & 0.067 & 0.241 \\
    Nishino \textit{et al.} \cite{nishino-etal-2022-factual} & JLiverCT & - & - & - & 0.466 & 0.592  \\
\bottomrule
\end{tabular}
}
\vspace{-0.2cm}
\end{table}

\smallbreak\noindent\textbf{Comparison across Different Approach Categories}
reports
the average performance of visual-only, textual-only, and cross-modal approaches on two main benchmark datasets (i.e., IU X-Ray \cite{iu-xray} and MIMIC-CXR \cite{mimic-cxr}), where we observe that the cross-modal approaches obtain the best performance on average, and the average score of the textual-only approach shows the second best, while the visual-only approach obtains the least performance.
There are several conclusions drawn with respect to different categories of RRG studies.
First, cross-modal approaches \cite{jing-etal-2018-automatic, jing-etal-2019-show, nishino-etal-2020-reinforcement, chen-etal-2020-generating, liu-etal-2021-contrastive, cvpr-2021-exploring, iccv-2021-visual-textual, chen-etal-2021-cross-modal, cvpr-2021-self-boosting, liu-etal-2021-competence, delbrouck-etal-2022-improving, qin-song-2022-reinforced, cvpr-2023-kiut, cvpr-2023-metransformer} reveal promising performance for RRG, indicating that the pivotal roles of both radiographs and reports for RRG, especially the image-text mapping between them, where Figure \ref{fig: case-study} presents the visualization of the cross-modal mapping between specific regions and medical terms in the generated report.
Second, textual-only approaches \cite{aaai-2019-knowledge, aaai-2020-when, cvpr-2021-exploring, yan-etal-2021-weakly-supervised, you-etal-2022-jpg, nishino-etal-2022-factual, kale-etal-2023-kgvl, cvpr-2023-dynamic, hou-etal-2023-organ, kale-etal-2023-replace} assist the report generation process with extra medical information (e.g., medical terms and related reports), which provide different granular levels of guidance for RRG, so that descriptive texts in the generated report are well-aligned with the corresponding terms in the gold standard report as is shown in Figure \ref{fig: case-study}.
Compared to cross-modal approaches, textual-only approaches mainly facilitate RRG from the perspective of the textual modality, and fail to address the enhancement for processing radiographs, thereby obtaining fewer improvements compared to cross-modal approaches.
Third, visual-only approaches \cite{cov-ctr, cvpr-2023-interactive} obtain the least performance among all method categories for RRG, which indicates that extracting elaborated features from radiographs is a challenging task that is hard to effectively address, where more related studies in the future are expected to facilitate RRG from such perspective.

\section{Challenges and Future Directions} \label{sec: limitations_and_future_directions}

As an interwoven research direction, RRG is intrinsically challenging that integrates the difficulties from both areas of AI and clinical medicine, which requires studies to consider the characteristics from AI models, algorithms as well as task-specific features in the medical domain, e.g., specific neural model architecture, different types of medical images, and the contextual specialties in reports, etc.
Although the methodology review and experimental analysis of existing studies have demonstrated
significant prosperity of RRG research 
over the past few years,
%
many open challenges and opportunities still stand, and being worthy of, for putting sustainable attention upon,
where data resources, model design, and evaluation protocols are three major directions,
in our opinion, to be thoroughly investigated and developed
to meet the requirements of real-world applications among many other important directions.
We elaborate the details of our analysis on them in the following subsections.

\subsection{Data Resources}


Data resources, which directly influence the performance of the RRG models by their participating in the training and evaluation processes, play an essential role in RRG research.
As mentioned in previous sections, existing approaches mainly conduct their experiments on benchmark datasets,
which have the potential to be improved in scale, where their data instances are uniformed in types and lack diversity, making it difficult for the trained models to meet various real-world clinical needs.
Thus, collecting and utilizing a wider range of data for this task is an anticipated direction for future research, which is elaborated as follows.

%
First, existing publicly available RRG datasets have the potential to be further improved in data quantity, which allows RRG approaches to achieve performance improvements through additional data.
%
Although there are massive amounts of unlabeled data in radiograph datasets (e.g., NIH chest X-ray dataset \cite{chexray8-dataset}) and medical corpora (e.g., PubMed\footnote{https://pubmed.ncbi.nlm.nih.gov/download/}), the supervised learninng paradigm of existing models are not able to utilize them for RRG owing to the fact of no radiograph-report pairs.
Meanwhile, collecting datasets for RRG studies is normally resource-consuming and requires extensive data cleaning efforts to ensure data quality, which still cannot be fully automated and thus demands a considerable amount of human efforts.
Therefore, to address the data quantity problem, approaches that utilize data with more efficiency are expected to improve RRG with data augmentation and unsupervised learning.

Second, existing datasets only support one type image-text mapping, which mainly consists of X-ray images and their corresponding reports.
As a result, X-ray images provide limited information to physicians in clinical practice, and struggle to present comprehensive knowledge from additional aspects, e.g., the health status of particular organs, temporal information, and other views of patients.
To address such limitations, other types of medical imaging are considered to provide enriched information for report generation.
For example, magnetic resonance imaging (MRI) is able to present specific states of organs, ultrasound imaging demonstrates the temporal development of a patient's health condition, and 3D ultrasound imaging offers a holistic view of the patient to radiologists.
Also, from the language perspective, existing radiology reports in public RRG datasets, especially the benchmark ones, are mainly written in English, whereas reports in other languages are also expected to promote the development of RRG research in different districts.
Consequently, approaches that are capable of processing diverse radiographs and generating reports in multiple languages are expected to be studied.

Third, widely used datasets mainly focus on chest radiographs and reports, where there is still high demand for radiographs and reports for other body regions, particularly for pivotal parts  (e.g., brains, bone joints), thereby resulting in difficulties in applying prevailing RRG approaches to clinical practice.
Particularly in these regions, the imaging form and medical terminology in the corresponding reports have significant differences compared to the ones of chest X-ray images.
Future research on more regions is thus highly expected to expand the application range of existing studies.
%

\subsection{Model Design}

As reviewed in previous sections,
current deep learning-based RRG approaches are mainly restricted in using few well-established frameworks (e.g., CNN for visual encoding and LSTM/Transfomer for text decoding),
which are, to some extent, highly integrated and not easy to be replaced and completely overturned,
so that lead to the challenging fact that many studies are limited to incremental innovations in their methodologies.
However, with the fast growing of AI, there are new possibilities in applying new techniques and even new paradigm to RRG, therefore many expectations are drawn in terms of model design for RRG.

Despite of using new models, the first one needs specific attention is the interpretability of RRG approaches.
Although promising performance has been obtained on multiple datasets, existing studies still lack of reasonable explanation for how their model is designed and why they work.
%
In trying to interpret the internal mechanism of their approaches, most studies \cite{jing-etal-2018-automatic, iccv-2021-visual-textual, chen-etal-2020-generating, chen-etal-2021-cross-modal, qin-song-2022-reinforced, cvpr-2023-metransformer} analyze the cross-modal correlation between input radiographs and generated reports through heat map visualizations. 
However, such solutions are still unable or ambiguous to explicitly demonstrate the 
underlying mechanisms and reasons behind that contribute to RRG
e.g., clinical theories, medical knowledge, diagnostic results, etc., where more explainable studies are expected in future RRG studies, especially in designing a model that is not performing in a black box.

From the aspect of radiograph processing, although existing studies employ different features for RRG, they struggle to represent radiographs in a detailed manner, leading to inaccuracies in generated reports. 
%
Even through some approaches adopt regional visual features \cite{cvpr-2021-self-boosting, cvpr-2023-interactive} to represent radiographs, the regions obtained by these approaches are still relatively coarse with a high degree of overlapping, leading to less accurate representations of radiographs and limited visual information for the report generation process.
%
To address such limitations, scaled-up image segmentation models (e.g., segment anything (SAM) \cite{Kirillov_2023_ICCV} and its variants for medical images \cite{SAM4MIS, samed, anand2023oneshot, ranem2023exploring, Pandey_2023_ICCV, wang2023samocta, zhang2023segment, adapting-sam-for-oct}), have shown potentials in extracting more fine-grained visual representations from radiographs, so that such models or similar designs should be able to provide more detailed visual information to enhance RRG.

For the generation process, the overwhelm majority of existing approaches 
%
use autoregressive models (i.e., LSTM/Transformer), and train them on public benchmark datasets from scratch.
%
Although such approaches demonstrate outstanding performance,
they are susceptible to the error propagation problem when incorrect texts are half-way produced during the report generation process.
Therefore, non-autoregressive (non-AR) models (e.g., diffusion model \cite{ho2020denoising}) are in the consideration for this task to offer an alternative paradigm through generating all words in a parallel manner, which has already demonstrated its effectiveness in other text generation tasks, e.g., sequence-to-sequence text generation \cite{gong2023diffuseq} and image captioning \cite{luo2022semantic, chen2023analog}.
As a most important notice, by witnessing current circumstance on the development of large language models (LLMs) \cite{touvron2023llama, touvron2023llama2}, one should admit that LLMs have demonstrated their outstanding generation and in-context learning ability in few shot settings, providing a promising potential solution to generate elaborated radiology reports.
Also, the in-context learning of LLM suggests possible variants of the RRG task, e.g., medical visual question answering and interactive report generation.
Furthermore, recent vision-language models (VLMs) \cite{zhu2023minigpt4, liu2023llava, ye2023mplugowl, xu2023mplug2} achieve impressive understanding and generation abilities through aligning vision backbone models (e.g., ViT) and LLMs, therefore showing their strength to enhance the radiograph-report alignment for RRG.
For example, recent studies such as XrayGPT \cite{thawkar2023xraygpt} adopt LLMs for radiology report summarization that generates a summary according to the input radiograph and the findings section of the report, which already proves the trend of introducing LLM for RRG.
However,
one possible concern for applying LLM to RRG is the domain adaptation problem, which requires adapting the knowledge of LLM from the general domain to the particular medical domain, especially with a limited amount of medical data.
A Challenging mission worth dedicated research is thus 
how to effectively perform such adaptation,
since domain mismatch is a difficult gap to overcome in applying LLMs to different real-world scenarios even though they possess strong few-shot learning ability.
%
%
%
%

\subsection{Evaluation Protocols}

Existing metrics for RRG evaluation generally follow the conventional settings of the AI field with using several routine measurements,
in either using a reference report or dedicated labels to compare with the results from different models on the textual side.
One significant limitation of doing so is the lack of cross-modal consideration with specific assessment on how good that particular regions in a radiograph and the texts in the report are associated with each other.
Although attention heat map is widely adopted to present the alignment between specific regions and medical labels in attention- and Transformer-based approaches \cite{jing-etal-2018-automatic, iccv-2021-visual-textual, chen-etal-2020-generating, chen-etal-2021-cross-modal, qin-song-2022-reinforced, cvpr-2023-metransformer}, there still requires quantitative evaluation to demonstrate such cross-modal connections.
In other multi-modal tasks, e.g., image captioning, there are already some automatic metrics to assess the cross-modal correlation between generated texts and the input images.
For example, CLIPScore \cite{hessel-etal-2021-clipscore} is used to evaluate the alignment between texts and images considering the distance of their features in the same semantic space, which offers a reference to measure the cross-modal relations for RRG.
From another aspect, existing metrics require references to measure the quality of the generated reports, yet it is generally hard to have such references standby in real-world applications.
Therefore, activity involving human evaluation has provided a practical option to measure the quality of radiology reports as well as cross-modal alignment between particular regions in radiographs and the texts in reports, and demonstrated significant potential in a series of scenarios such as telemedicine consultation.
For example, A/B testing for physicians' references on generated reports is a promising solution that presents radiologists with comparisons by different RRG approaches as well as providing feedbacks the reports' quality and clinical validity according to human preferences.

\section{Conclusion}

RRG plays a pivotal role in applying AI techniques to the clinical medicine field.
In panoramically investigating the development of recent RRG research, in this paper,
we propose to categorize existing RRG studies into three groups based on their enhancing modalities, namely visual-only, textual-only, and cross-modal approaches.
Then, we summarize the evaluation standards for RRG, consisting of different datasets and evaluation metrics.
Afterwards, we introduce the experiment settings of different approaches and their results, as well as discuss their impacts based on their model architecture and approach categories.
Finally, we present challenges and future directions according to technological advancements and the demands of real-world requirements.
%
Although early studies \cite{pavlopoulos-etal-2019-survey, rrg-survey-2020, rrg-survey-2022} that also survey RRG research from perspectives of different aspects, e.g., its domain and techniques,
which mainly categorize RRG studies by comparing RRG with other similar tasks (e.g. image captioning) and put emphasis on particular components (e.g., activation functions) in their models, rather than focusing on the modality differences in methodology settings.
Compared to them, this paper offers an updated and systematic review of RRG by focusing on deep learning-based approaches, which is expected to serve as a comprehensive half-decade summary with helpful insights to guide future research in this particular area.

\appendices

\ifCLASSOPTIONcaptionsoff
  \newpage
\fi

\bibliographystyle{IEEEtran}
\bibliography{reference}

\end{document}